\documentclass{clv3}
\usepackage{hyperref}
\usepackage{xcolor}
\usepackage{standalone}
\usepackage{multirow}
\usepackage{booktabs}
\usepackage[table]{colortbl}
\usepackage{rotating}
\usepackage{tabularx}
\usepackage{expex}
\usepackage{paralist}
\usepackage{longtable}
\usepackage{subcaption}
\usepackage{pgfplots}
\usepackage{color}
\usepackage[utf8]{inputenc}

\usepackage{ragged2e}
\usepackage[colorinlistoftodos]{todonotes}

\newcommand{\rev}[2]{#2}
\newcommand{\secrev}[2]{#2}
\newcommand{\ikrev}[2]{\textcolor{violet}{#2}}

\definecolor{darkblue}{rgb}{0, 0, 0.5}
\hypersetup{colorlinks=true,citecolor=darkblue, linkcolor=darkblue, urlcolor=darkblue}

\dochead{LINSPECTOR}

\runningtitle{Multilingual Probing Tasks for Word Representations}

\runningauthor{Gözde Gül Şahin}

\begin{document}

\title{Multilingual Probing Tasks for Word Representations}

\author{Gözde Gül Şahin\thanks{Research Training Group AIPHES and UKP Lab, Department of Computer Science, Technische Universität Darmstadt, Darmstadt, Germany. E-mail: sahin@ukp.informatik.tu-darmstadt.de.}}
\affil{AIPHES and UKP Lab / TU Darmstadt}

\author{Clara Vania\thanks{Work done while Clara Vania was a PhD student at the ILCC / University of Edinburgh, E-mail: c.vania@nyu.edu}}
\affil{New York University}

\author{Ilia Kuznetsov}
\affil{AIPHES and UKP Lab / TU Darmstadt}

\author{Iryna Gurevych}
\affil{AIPHES and UKP Lab / TU Darmstadt}

\maketitle

\begin{abstract}
    Despite an ever growing number of word representation models introduced for a large number of languages, there is a lack of a standardized technique to provide insights into what is captured by these models.
    Such insights would help the community to get an estimate of the downstream task performance, as well as to design more informed neural architectures, while avoiding extensive experimentation which requires substantial computational resources not all researchers have access to. 
    A recent development in NLP is to use simple classification tasks, also called \textbf{probing tasks}, that test for a single linguistic feature such as part-of-speech. 
	Existing studies mostly focus on exploring the \rev{information encoded by the sentence-level representations for}{linguistic information encoded by the continuous representations of} English \rev{}{text}. However, from a typological perspective the morphologically poor English is rather an outlier: the information encoded by the word order and function words in English is often stored on a subword, morphological level in other languages.
    To address this, we introduce 15 \rev{word}{type}-level probing tasks such as case marking, possession, word length, morphological tag count and pseudoword identification for 24 languages.
    We present a reusable methodology for creation and evaluation of such tests in a multilingual setting, which is challenging due to lack of resources, lower quality of tools and differences among languages. 
    We then present experiments on several \rev{state of the art}{diverse multilingual} word embedding models, in which we relate the probing task performance for a diverse set of languages to a range of five classic NLP tasks: POS-tagging, dependency parsing, semantic role labeling, named entity recognition and natural language inference.
    We find that a number of probing tests have significantly high positive correlation to the downstream tasks, especially for morphologically rich languages. 
    We show that our tests can be used to explore word embeddings or black-box neural models for linguistic cues in a multilingual setting. We release the probing datasets and the evaluation suite \rev{with}{LINSPECTOR with} \url{https://github.com/UKPLab/linspector}. 
\end{abstract}

\section{Introduction}
\label{sec:intro}



The field of natural language processing (NLP) has seen great development after replacing the traditional discrete word representations with continuous ones. Representing text with dense, low-dimensional vectors - or \emph{embeddings} - has become the \emph{de facto} approach, since these representations can encode complex relationships between the units of language and can be learned from unlabeled data, thus eliminating the need for expensive manual feature engineering. The initial success of dense representations in NLP applications has led to the development of a multitude of embedding models, which differ in terms of design objective \textit{(monolingual~\citep{Mikolov:2013:DRW:2999792.2999959}, cross-lingual~\citep{ruder2017survey}, contextualized~\citep{peters:NAACL2018}, retrofitted~\citep{faruqui2015retrofitting}, multi-sense~\citep{PilehvarCNC17}, cross-domain~\citep{YangLZ17}, dependency-based~\citep{LevyG14})}, encoding architecture, \textit{(convolution~\citep{kim2015}, linear vector operations~\citep{bojanowski:TACL2017}, bidirectional LSTM~\citep{ling2015})}, as well as in terms of the target units \textit{(words, characters, character n-grams, morphemes, phonemes)}. 


While offering substantial benefits over the traditional feature-based representations of language, the performance of unsupervised embeddings may differ considerably depending on the language and the task. 
For instance, early embedding models such as word2vec~\citep{Mikolov:2013:DRW:2999792.2999959} and GloVe~\citep{glove:14} have been shown to suffer from out-of-vocabulary (OOV) issues for agglutinative languages like Turkish and Finnish~\citep{sahin:acl18}, while performing relatively well on analytic and fusional languages like English. Furthermore, there is no guarantee that a representation well-suited for some task would score similarly well at other tasks even for the same language due to the differences in the information required to solve the tasks, as demonstrated in~\citet{rogers2018s}. 


Given the variety of word representations and parameter options, searching for the right word \rev{embedding/encoder}{representation} model for a specific language and a certain task is not trivial. Scanning the large parameter space may be extremely time consuming and computationally expensive, which poses significant challenges, especially in the lower-resource non-English academic NLP communities. To simplify the search for a good representation, and estimate the ``quality'' of the representations, intrinsic evaluation via similarity and analogy tasks has been proposed. While these tasks seem to be intuitive, there are concerns regarding their consistency and correlation with downstream task performance \citep{linzen2016issues,SchnabelLMJ15}. Furthermore, such evaluation requires manually created test sets and these are usually only available for a small number of languages. Another \rev{popular technique} option to assess the quality of word representation is through extrinsic evaluation, where the word vectors are used directly in downstream tasks, such as machine translation (MT)~\citep{ataman2018}, semantic role labeling (SRL)~\citep{sahin:acl18} or language modeling (LM)~\citep{vania2017}. Although this method provides more insightful information about the end task performance, it requires expensive human annotations, computational resources and the results are sensitive to hyperparameter choice.


To address the aforementioned problems, a few studies have introduced the idea of \emph{probing tasks} \citep{kohn2016evaluating,shi-etAl:ACL2016,adi2017fine,Veldhoen2016DiagnosticCR,senteval18}; which are a set of \rev{multi-label}{multi-class} classification problems that probe a learned word vector for a specific linguistic property, such as part-of-speech (POS), semantic, or morphological tag\footnote{We use the terms probing tasks and probing tests interchangeably throughout the paper.}. 
Probing tasks have gained a lot of attention~(\citet{belinkov:acl2017,bisazza:emnlp2018}, among others) due to their simplicity, low computational cost, and ability to provide some insights \rev{into}{regarding} the linguistic properties \rev{of the word}{that are captured by the learned} representations. 

The majority of the probing tests proposed so far are mostly designed for \textbf{English language only}, and operate on the \textbf{sentence-level} (e.g. \emph{tree depth}, \emph{word count}, \emph{top constituent} by \citet{senteval18}). Although sentence-level probing may provide valuable insights for English sentence-level representations, \rev{}{we hypothesize that they would not be similarly beneficial in a multilingual setup for several reasons.} The first reason is that the information encoded by the word order and function words in English is encoded at the morphological, subword level information in many other languages. Consider the Turkish word \emph{katılamayanlardan}, that means ``he/she is one of the folks who can not participate''. In morphologically complex languages like Turkish, single tokens might already communicate a lot of information such as event, its participants, tense, person, number, polarity. In analytic languages, this information would be encoded as a multi-word clause. The second reason is the confusion of the signals: as pointed out by \citet{tenney2018what}, sometimes ``operating on full sentence encodings introduces confounds into the analysis, since sentence representation models must pool word representations over the entire sequence''. \rev{}{Furthermore, we argue that such tests would carry over the statistics of the data they originate from, introducing undesired biases such as domain and majority bias. In order to address the aforementioned issues, we introduce context independent, dictionary-based \textbf{type-level} probing tasks that operate on word-level and do not contain domain or majority biases. To investigate the limitations and strengths of the proposed type-level tasks, we introduce and investigate another set of similar, but context dependent, treebank-based and thereby potentially biased \textbf{token-level} tests.}

In this work,
\begin{itemize}
\item We extend the line of work by~\citet{senteval18} and \citet{tenney2018what} and introduce \textbf{15 type-level} probing tasks for \textbf{24 languages} by taking language properties into account. Our probing tasks cover a range of features: from superficial ones such as word length, to morphosyntactic features such as case marker, gender, and number; and psycholinguistic ones like pseudowords (artificial words that are phonologically well-formed but have no meaning). Although languages share a large set of common probing tasks, each has a list of its own, e.g., Russian and Spanish are probed for gender, while Turkish is probed for polarity and possession; 
\item We introduce a reusable, systematic methodology for creation and evaluation of such tests by utilizing the existing resources such as UniMorph~\citep{sylakGlassmanK15,sylak2016composition,kirov-etal-2018-unimorph}, Wikipedia and Wuggy~\citep{keuleers2010wuggy};
\item We then use the proposed probing tasks to evaluate \rev{the most commonly used multilingual word embedding models}{a set of diverse multilingual embedding models} and to diagnose a neural end-to-end semantic role labeling model as a case study. We statistically assess the correlation between probing and downstream task performance for a variety of downstream tasks (POS tagging, dependency parsing (DEP), semantic role labeling (SRL), named entity recognition (NER) and natural language inference (NLI)) for a set of typologically diverse languages and find that a number of probing tests have significantly high positive correlation to a number of syntactic and semantic downstream tasks, especially for morphologically rich languages;

\item \rev{}{We introduce a set of comparable \textbf{token-level} probing tasks that additionally employs the context of the token. We analyze the type- and token-level probing tasks through a series of intrinsic and diagnostic experiments and show that they are similar with some exceptions: token-level tasks may be influenced by \emph{domain and majority class biases}, while type-level tasks may suffer in case of \emph{lack of lexical diversity} and high \emph{ambiguity ratios};}

\item \rev{}{We provide comprehensive discussions for the intrinsic and extrinsic experimental results along with diagnostic and correlation study. We show that numerous factors except from the neural architectures play role on the results such as out-of-vocabulary rates, domain similarity, statistics of both datasets (e.g., ambiguity, size), training corpora for the embeddings; as well as typology, language family, paradigm size and morphological irregularity.}

\item We release the LINSPECTOR framework that contains the probing datasets along with the intrinsic and extrinsic evaluation suite: \url{https://github.com/UKPLab/linspector}.
\end{itemize}
We believe our evaluation suite together with probing datasets could be of great use for comparing various multilingual word representations such as automatically created cross-lingual embeddings; exploring the linguistic features captured by word encoding layers of black-box neural models; systematic searching of model or architecture parameters \rev{}{by evaluating the models with different architectures and parameters on the proposed probing tasks}; or comparing transfer learning techniques \rev{by probing for the transferred linguistic knowledge from high-resource to low-resource languages}{i.e. by evaluating a set of cross-lingual embeddings that are transfered or learned using different transfer learning techniques} \rev{}{, on the proposed language-specific probing task set}.

\section{Related Work on Word Representation Evaluation}
\rev{}{
We begin with a review of related work on word representation evaluation. We divide the current evaluation schemes for word representations into two main categories: (1) \textit{intrinsic}, when vectors are evaluated on a variety of benchmarks and (2) \textit{extrinsic}, when they are evaluated on downstream NLP tasks. 
}

\subsection{Intrinsic evaluation} 

\rev{A standard way to assess and compare continuous word representations is the pairwise similarity benchmarking: a set of word pairs is manually annotated with respect to some notion of similarity, and the similarity scores produced by the word embedding model are compared to the human annotators in terms of correlation. Most commonly used word similarity datasets for English are WordSim-353 \citep{FinkelsteinGMRSWR01}, MC \citep{miller&charles}, 
SCWS \citep{Huang:ACL2012}, rare words dataset (RW) \citep{Luong-etal:conll13:morpho}, MEN \citep{BruniBBT12}, and SimLex-999 \citep{HillRK15}. The size of these datasets ranges from 30 to 999 word pairs with a focus on English.}
{A standard approach to evaluate continuous word representations is by testing them on a variety of benchmarks which measures some linguistic properties of the word. These similarity benchmarks typically consist of a set of words or word pairs that are manually annotated for some notion of relatedness (e.g., semantic, syntactic, topical, etc.). For English, some of the widely used similarity benchmarks are WordSim-353 \citep{FinkelsteinGMRSWR01}, MC \citep{miller&charles}, RG \citep{Rubenstein:1965:CCS:365628.365657}, SCWS \citep{Huang:ACL2012}, rare words dataset (RW) \citep{Luong-etal:conll13:morpho}, MEN \citep{BruniBBT12}, and SimLex-999 \citep{HillRK15}. While these benchmarks have shown to be useful for evaluating English word representations, only very few word similarity datasets exist in other languages. \citet{Leviant2015SeparatedBA} collected human-assessed translations of WordSim-353 and SimLex-999 on three languages, Italian, German and Russian.\footnote{http://leviants.com/ira.leviant/MultilingualVSMdata.html} For SemEval 2017 shared task, \citet{camachocollados-EtAl:2017:SemEval} introduced manually curated word-similarity datasets for English, Farsi, German, Italian, and Spanish.
}

Another popular benchmark for evaluating word representations is the word analogy test. This test was specifically introduced by \citet{Mikolov13} to evaluate word vectors trained using neural models. The main goal is to determine how syntactic and semantic relationships between words are reflected in the continuous space. Given a pair of words, \textit{man} and \textit{woman}, the task is to find a target word which shares the same relation with a given source word. For example, given a word \textit{king}, one expected target word would be \textit{queen}. The analogy task has gained a lot of attention mainly because it demonstrates how ``linguistic regularities'' are captured by word representation models. The analogy dataset of \citet{Mikolov13} consists of 14 categories covering both syntactic and semantic regularities. Although analogy test has become a standard evaluation benchmark, \citet{rogers-etAl:SEM2017} and \citet{linzen2016issues} identified certain theoretical and practical drawbacks of this approach, which are mostly related to the consistency of the vector offset and the structure of the vector space model. Pairwise similarity benchmarks and word analogy tasks only offer a first approximation of the word embedding properties and provide limited insights into the downstream task performance. To address this limitation, \citet{tsvetkov2015evaluation} introduced QVEC, an intrinsic word evaluation method which aligns word vector representations with hand-crafted features extracted from lexical resources, focusing on the semantic content. They showed that their evaluation score correlates strongly with performance in downstream tasks.

More recently, \citet{rogers2018s} proposed a comprehensive list of scores, so-called linguistic diagnostics factors, and analyzed their relation to a set of downstream tasks such as chunking, named entity recognition (NER), sentiment classification, using word2vec \citep{Mikolov13} and GloVe \citep{glove:14} word representations. They extend the traditional intrinsic evaluation (word similarity and analogy) with semantics extracted from existing resources such as WordNet, and basic morphological information like shared lemma and affixes. Their findings support the previous studies that observe low correlation between word similarity/analogy and sequence-labeling downstream task performance. In addition, they observe high correlation between morphology-level intrinsic tests with such downstream tasks even for English - one of the morphologically poorest languages. Unlike probing studies that train classifiers, they rely on nearest neighbor relation as a proxy to predict the performance of word vectors similar to early word analogy works.  

\subsection{Extrinsic evaluation}

In general, evaluating word vectors on downstream NLP tasks is more challenging because of the time and resources needed for the implementation. The two most common approaches are to test a single representation model on several downstream tasks \citep{ling2015,glove:14,bojanowski:TACL2017}, or to test a number of representation models on a single task \citep{vania2017,ataman2018,sahin:acl18,gerz-etAl:TACL2018}. For a more general extrinsic evaluation, we note the work of \citet{NayakAM16}, which introduces an evaluation suite of six downstream tasks: two tasks to assess the syntactic properties of the representations and four tasks to assess the semantic properties. \rev{}{Since this type of evaluation is typically task-specific, it can be conducted in multilingual settings. However, training a range of task-specific multilingual models might require significant resources, i.e., training time and computational power. Apart from that, differences in the exact task formulation and the underlying datasets among languages might influence the evaluation results.}

\subsection{Evaluation via probing task}

The rise of deep learning based methods in NLP has stimulated research on the interpretability of the neural models. In particular, several recent studies analyze representations generated by neural models to get insights on what kind of linguistic information is learned by the models. Interpretability studies have been one of the emerging trends in NLP as hinted by the on-going Representation Evaluation (RepEval)~\citep{nangia2017repeval} and BlackBoxNLP Workshop series~\citep{W18-5400} organized in popular conference venues. The most common approach is to associate some linguistic properties such as POS, \rev{}{morphological, or semantic properties} with \rev{}{specific} representations from a \rev{}{trained} model \rev{like the encoding layer or the output of the activation layer}{(hidden states or activation layer)}. This method, which is called \emph{probing task} or \emph{diagnostic classifier} \citep{shi-etAl:ACL2016,adi2017fine,Veldhoen2016DiagnosticCR}, uses representations generated from a fully-trained model with frozen weights to train a \rev{linear}classifier predicting a particular linguistic property. The performance of this classifier is then used to measure how well the model has ``learned'' this particular property. A similar study has been conducted by ~\citet{kohn-2015-whats} which proposed training such classifiers for predicting syntactic features such as gender and tense, extracted from annotated dependency treebanks. Due to unavailability of subword or contextualized embeddings at that time, the author only experiment with static word-level embeddings (word2vec, GloVe, and embeddings derived from Brown clusters) and find that \rev{they all perform similarly}{they are suprisingly able to capture linguistic properties, in particular for POS information}. The study assumes that the performance of this targeted word feature classifiers would be directly related to the parser performance, \rev{}{which is later tested empirically with diagnostic classifiers on syntactic parsers for simple linguistic properties such as tense and number~\cite{kohn2016evaluating}}. \rev{}{Although the syntax-based classifiers in \namecite{kohn2016evaluating} are conceptually similar to our single feature probing tasks, there are several differences. First, the training instances are created from an annotated treebank including the ambiguous words; which may introduce domain, annotator and majority class bias unlike ours; and lead to inconsistent results due to unresolved ambiguity. In addition, the study is limited to the following tests: case, gender, tense and number; and to syntactic parsing as the downstream task. Finally, it has used only three word embedding models, which can be considered too small to draw conclusions; and too similar, as their training objectives and training units are similar.} 

\rev{}{\citet{qian-etal-2016-investigating} investigate the effects of word inflection and typological diversity in word representation learning. They observe that language typology (word order or morphological complexity) influences how linguistic information is encoded in the representations. They also compare a standard character-level auto-encoder model to a word-level model (word2vec Skip-Gram) and find that character-level models are better at capturing morphosyntactic information. Their study highlights the importance of utilizing word form information as well as language typology.}

Recent works on probing have focused on analyzing the representations learned when training for specific downstream tasks, such as machine translation \citep{shi-etAl:ACL2016,belinkov:acl2017,bisazza:emnlp2018} or dependency parsing \citep{vania:EMNLP2018}. While this approach allows probing for multilingual data, it is still task-specific and might require expensive computation for model training (e.g., machine translation typically needs a large amount parallel data for training). \rev{as well as on the sentence-level representations evaluation \citep{senteval18,tenney2018what}}{For a more general evaluation, \citet{senteval18} and \citet{tenney2018what} each introduced a broad coverage evaluation suite to analyze representations on the sentence level with focus on English}. We build our methodology upon these recent works. \rev{}{However, }unlike their methods, our evaluation suite is multilingual and takes language-specific features into account. Moreover our tests are \rev{word}{type}-level, rather than sentence~\citep{senteval18} or sub-sentence level \citep{tenney2018what}. 

\rev{}{Finally, \citet{belinkov-glass-2019-analysis} recently surveyed various analysis methods in NLP and mention three important aspects for model analysis: (1) the methods (classifiers, correlations, or similarity), (2) the linguistic phenomena (sentence length, word order, syntactic, or semantic information, etc), and (3) the neural network components (embeddings or hidden states). They have also provided a non-exhaustive list of previous work which use probing task (classifier) method for analyzing representations, including word representations. For a more comprehensive list of studies on what linguistic information is captured in neural networks we refer the readers to \citet{belinkov-glass-2019-analysis}}


\section{Probing Tasks}
	

With our probing tasks we aim to cover the properties ranging from shallow e.g. \emph{word length}~\citep{senteval18}, to deeper ones e.g. distinguishing \emph{pseudowords} from in-vocabulary words. First, we probe for morphosyntactic and morphosemantic features such as \emph{case marking}, \emph{gender}, \emph{tense} and \emph{number}. Most probing tasks are defined for all languages, such as \emph{POS} and \emph{number}, however, some features are only defined for a subset of languages, e.g., \emph{polarity} for Portuguese and Turkish, \emph{gender} for Arabic and Russian. To maintain consistency, we base the majority of our tasks on the universal grammatical classes introduced by UniMorph project \citep{sylakGlassmanK15}. Second, we propose tasks to evaluate a more general syntactic/semantic capability of the model such as predicting the number of \emph{morphological tags}, detecting the shared and \emph{odd} linguistic feature between two word forms. Finally, inspired by cognitive linguistics, we assess the ability of the embedding models to detect pseudowords, i.e., words that are phonetically similar to an existing word but have no meaning. The conceptual definitions of our probing tests are given in Section~\ref{ssect:tesdefs}, while Section~\ref{ssect:testcreation} describes the specific implementation of the probing tests used in this work.

\subsection{Task Definitions}
\label{ssect:tesdefs}

\paragraph{Case Marking} A substantial number of languages express the syntactic and semantic relationship between the nominal constituents and the verbs via morphological case markers. \namecite{casemarking} reports that 161 out of 261 languages have at least two case markers as shown in Table~\ref{tab:numcase}. 
\begin{table}[!ht]
      \begin{tabular}{lll}  
      \textbf{\#Case Categories} & \textbf{\#Languages} & \textbf{Example} \\
      \hline 
      0 & 100 & English, Spanish \\ 
      2 & 23 & Romanian, Persian \\ 
      3 & 9 & Greek \\ 
      4 & 9 & Icelandic, German, Albanian \\          
      5 & 12 & Armenian, Serbo-Croatian, Latvian \\ 
      6-7 & 37 & Turkish, Polish, Russian, Georgian \\ 
      8-9 & 23 & Japanese \\
      10 or more & 24 & Estonian, Finnish, Basque \\ 
      \end{tabular}
  \caption{Languages with case marking}
  \label{tab:numcase}
\end{table} 
Although cases may undertake different roles among languages, a type of case marking, named as \textit{core}, \textit{non-local}, \textit{nuclear} or \textit{grammatical case}, is the most common. This category contains case markers that are used to mark the arguments of verbs such as subjects, objects and indirect objects~\cite{casemarking2,casemarking3}. In languages with rich case marking systems, case is also commonly used to mark roles such as ``location'' and ``instrument''. Below are examples of Russian and Turkish sentences that use \texttt{Acc} and \texttt{Inst} case markers to define the patient (object affected by the action) and the instrument.    
\pex
\a
\begingl
\gla Mark-$\emptyset$ razbi-l-$\emptyset$ okn-o molotk-om //
\glb Mark-\sc{Nom.Sg} break-\sc{Pst-Sg.M} window-\sc{Acc.Sg} hammer-\sc{Inst.Sg} //
\endgl
\a
\begingl
\gla Mark-$\emptyset$ pencere-yi çekiç-le kır-dı //
\glb Mark-\sc{Nom.Sg} window-\sc{Acc.Sg} hammer-\sc{Inst.Sg} break-\sc{Pst.3.Sg}//
\glft  \rev{}{‘Mark broke the window with a hammer.’}//
\endgl
\xe
The relation between case markers and NLP tasks such as semantic role labeling, dependency parsing and question answering have been heavily investigated and using case marking as feature has been shown beneficial for numerous languages and tasks~\cite{isguderA14,eryigitNO08}.

\paragraph{Gender} 
According to \namecite{gender}, more than half of the world languages do not have a gender system. Majority of the languages with a gender system such as Spanish, French, German, and Russian, define either two (feminine, masculine) or three (neutral) classes. Gender is a grammatical category and participates in agreement: if a language has a gender system, the gender of a noun or pronoun influences the form of its syntactic neighbors, which could be verb, adjective, determiner, numeral or a focus particle, depending on the language. \rev{}{Related to NLP tasks, \citet{hohensee-bender-2012} showed that agreement-based features including gender can improve the quality of dependency parsing for morphologically rich languages. \citet{Bengtson2008UnderstandingTV} demonstrate how gender can be used to improve co-reference resolution quality.}
In the Russian example sentence given below, the gender agreement between the subject, its adjective modifier and the verb is shown. 
\ex
\begingl
\gla Gosudarstvenn-aya duma sdela-l-a zayavlenie//
\glb State-\sc{Nom.Sg.F} parliament\sc{.F} make-\sc{Pst}-\sc{Sg.F} announcement//
\glft  \rev{}{‘The parliament made an announcement.’}//
\endgl
\xe
The agreement features such as gender and number are crucial for structured grammatical analysis such as dependency parsing, co-reference resolution, as well as for grammar checking and correction, and automatic essay evaluation.

\paragraph{Mood} 
Modality of the verb, i.e., the grammatical mood, is used to communicate the status of the proposition from the speaker's point of view. Some common mood categories are Indicative, Conditional, Subjunctive, Imperative-Jussive and Potential. Many languages mark the modality of the verb with morphological affixes. German and Russian example sentences with Imperative mood feature is given below. 
\pex
\a
\begingl
\gla Bring-e mir das Buch //
\glb Bring-\sc{2Sg.Imp} me the book //
\endgl
\a
\begingl
\gla Prines-i mne knigu //
\glb Bring-\sc{2Sg.Imp} me book //
\glft  \rev{}{‘Bring me the book.’}//
\endgl
\xe

\rev{Mood is an essential feature for dialogue systems, where the speaker's point of view is important in order to decide for the next action.} {Since Mood signals the factuality of the statement, it might be relevant for natural language inference and related tasks, as we demonstrate in Section \ref{ssect:correl}; the ability of the representation to encode imperative, in turn, could be essential for interpreting the user input in dialogue systems.}

\paragraph{Number} 
This feature is usually expressed by nouns, adjectives and verbs, and similar to gender, number is a common feature for agreement. The two most common values for gender is Singular and Plural, which often marked by morphological affixes.

\paragraph{POS} We use the following eight categories defined by the UniMorph Schema: nouns, adpositions, adjectives, verbs, masdars, participles, converbs, and adverbs. A more detailed information for each category can be found in \namecite{sylak2016composition}. POS has been one of the most prominent features of all high-level NLP tasks for decades. \rev{It is important to note that language-specific POS tags often encode additional information such as person and number. Throughout this work, however, we refer to coarse POS, a strictly grammatical category.}{Throughout this work, we will use the coarse POS categories, which are universal across languages.}

\paragraph{Person} We use the traditional six person categories that are commonly marked by morphological markers: 1st, 2nd and 3rd person either singular or plural. This feature has strategic importance for dependency parsing, co-reference resolution, as well as high-level tasks that involve natural language understanding such as conversational agents, question answering or multi-modal applications such as generating images from sentences. An example of using person agreement to improve dependency parsing quality is shown in \citet{hohensee-bender-2012}. \rev{}{Two Russian example sentences below demonstrate coordination between the personal pronoun and the verb, indicating a syntactic dependency between them.}

\pex
\a
\begingl
\gla Ja vizh-u ptitsu //
\glb I\sc{.1Sg} see-\sc{1Sg.Prs} bird//
\endgl
\a
\begingl
\gla On vid-it ptitsu //
\glb He\sc{.3Sg} see-\sc{3Sg.Prs} bird//
\glft  \rev{}{(a) ‘I see a bird.’ (b) ‘He sees a bird.’}//
\endgl
\xe

\paragraph{Polarity} 
Some languages mark the verbs \rev{}{with polarity} to indicate whether a statement is negative or positive. Generally, markers are used to specify the negative polarity, assuming the positive polarity by default. The verb ``go'' is marked with a negative marker in the Turkish sentence given below. Although this feature is not notably common across languages, it has immediate use cases such as sentiment analysis and natural language inference\rev{}{, similar to negation in English.}

\pex
\begingl
\gla Dün okul-a git-me-di-m //
\glb yesterday school-\sc{Dat.Sg} go-\sc{Neg}-\sc{Pst}-\sc{Sg}//
\glft  \rev{}{‘He/she didn't go to school yesterday.’}//
\endgl
\xe

\paragraph{Possession} 
Although majority of the languages use adjectives such as his/her/my to express possession, some languages such as Turkish and Arabic use morphological markers on the nouns. The number of values for the feature depends on the gender system of the language. For instance, while Arabic separately marks the possession by 3rd person singular for feminine and masculine, Turkish uses only one marker for the possession by the 3rd person singular.
\pex
\begingl
\gla Ayakkabı-(s)ı-(n)ı giy-ecek //
\glb shoe-\sc{Poss.3Sg}-\sc{Acc} wear-\sc{3Sg.Fut}//
\glft  \rev{}{‘He/she will wear his/her shoes.’}//
\endgl
\xe
An example sentence in Turkish with ``he/she will wear his/her shoes'' is given above. As can be seen, possession implicitly acts as an agreement feature, i.e., possession of the object and person of the verb must match. 

\paragraph{Tense} 
We use the simplified universal definition of tense, which is encoding of the event time. Similar to previous categories, we only account for the categories and the languages that have morphological markers for tense. The most common values for tense across languages in our dataset are: Past, Present and Future. Russian and German examples with Past tense marking are given below for reference.   
\pex
\a
\begingl
\gla On kupi-l-$\emptyset$ etot dom //
\glb He buy-\sc{Pst-Sg.M} this house //
\glft  \rev{}{‘He bought this house.’}//
\endgl
\a
\begingl
\gla Auf dem Tisch lag-$\emptyset$ ein Buch //
\glb On the\sc{.Dat} table lie.\sc{Pst}-\sc{Sg} a book //
\glft  \rev{}{‘There was a book on the table’}//
\endgl
\xe

Tense \rev{feature has a special importance for applications where resolving the event order is crucial.}{expresses the temporal order and the factuality of the events and states, and is therefore expected to contribute to inference and time-based NLP problems.}

\paragraph{Voice} 
This study is only concerned with frequently occurring Active and Passive voice features that have separate morphological markers in the verb. \rev{}{A synthetic German example using passive voice is given below. As shown, the semantic roles of \textit{he} (\texttt{Agent}\footnote{As per VerbNet 3.3, https://verbs.colorado.edu/verb-index/vn3.3/}) and \textit{house} (\texttt{Product}) are encoded differently depending on the voice of the main verb.}
\pex
\a
\begingl
\gla Er baut das Haus . //
\glb He.\sc{Nom} build.\sc{Act} the house.\sc{Acc} //
\glft  \rev{}{‘He builds the house’}//
\endgl
\a
\begingl
\gla Das Haus wird von ihm gebaut . //
\glb The house.\sc{Nom} is by he.\sc{Dat} build.\sc{Pass} //
\glft  \rev{}{‘The house is built by him’}//
\endgl
\xe

\rev{}{Since voice affects the encoding of the core semantic arguments, the ability of word embedding methods to represent voice information is expected to contribute to the dependency parsing and semantic role labeling and induction performance.}

\paragraph{Tag Count} 
We create a test that contains tuples of surface forms and number of morphological tags (annotated according to UniMorph schema) for the token. It can be considered a simplistic approximation of the morphological information encoded in a word and is expected to cover a mixture of the linguistic aspects outlined above. \rev{}{For instance the Turkish word ``deneyimlerine'' (to their/his/her/your experiences) annotated with \textsc{(N.Dat.Pl.Poss2Sg)} would have the tag count of 4, while ``deneyimler'' (experiences) annotated with \textsc{(N.Dat.Pl)} would have the count 3. It can also be associated with the model's capability of segmenting words into morphemes, \textit{i.e., morphological segmentation}, especially for agglutinative languages like Turkish where morpheme to meaning is a one-to-one mapping. Furthermore for fusional languages with one-to-many morpheme to meaning mapping, it can be associated with model's ability to learn such morphemes with multiple tags as in the Spanish word ``hablo''-\textsc{V.Ind.1Sg.Prs} (I speak), where ``o'' alone conveys the information about the mood, tense and the person.} 

\paragraph{Character Bin} 
Here we create a test set consisting of pairs of randomly picked surface forms and the number of unicode characters they contain. For convenience, we used bins instead of real values as in ~\namecite{senteval18}. \rev{Although this task is considered shallow, as we show later in the paper, it captures relevant information about morphology for agglutinative languages.}{The motivation behind this feature is to use number of characters as an approximation to number of morphological features, similar to previously motivated \textit{Tag Count} test. We hypothesize that this should be possible for agglutinative languages where there is one-to-one mapping between morpheme and meaning, unlike the mapping in fusional languages. \emph{Character Bin} can therefore be seen as a rough approximation of Tag Count with the advantage of being able} to expand this resource to even more languages since it does not require any morphological tag information.  

\paragraph{Pseudowords} 
Pseudowords or Nonwords are commonly used in psycholinguistics to study lexical choices or different aspects of language acquisition. There are various ways to generate pseudowords, e.g. randomly swapping two letters, randomly adding/deleting letters to/from a word; or concatenating high-frequency bi-grams or trigrams. These methods have limitations when it comes to multilingual studies such as computational time, availability of resources and researcher's bias as explained in details by~\namecite{keuleers2010wuggy}. In this study, we use the ``Wuggy'' algorithm \cite{keuleers2010wuggy} which is the most commonly used and freely available system for multilingual pseudowords generation. It builds a grammar of bi-gram chains from the syllabified lexicon and generates all possible words with the grammar, both words and nonwords. It is available for German, Dutch, English, Basque, French, Spanish and Vietnamese by default, and has been extended for Turkish~\cite{erten2014turkish}. 
Some examples of generated pseudowords from our dataset are given in Table~\ref{tab:pseudo}.
\begin{table}[!ht]
\center
      \begin{tabular}{lc}  
      \textbf{Language} & \textbf{Pseudowords} \\
      \hline 
      English & atlinsive, delilottent, foiry  \\ 
      French & souvuille, faicha, blêlament \\
      Basque & zende, kontsiskio, anazkile, kaukasun, kaldretu \\
      Dutch & nerstbare, openkialig, inwrannees, tedenjaaigige, wuitje \\
      Serbian & aćejujelu, benkrilno, knjivule, haknjskim, znamaketi \\
      German & Anstiffung, hefumtechen, Schlauben, Scheckmal, spüßten \\
      Spanish & vuera, espisia, supencinzado, lungar, disciscir \\
      Turkish & ular, pesteklelik, çanar, tatsazı, yalsanla \\
      \end{tabular}
  \caption{Examples of generated pseudowords}
  \label{tab:pseudo}
\end{table}
\rev{}{Since the Wuggy algorithm can generate words that sound natural, this test can be used to distinguish subword-level models that can capture semantic-level information from the ones that remain on ortography-level.}
\paragraph{SameFeat}
We choose two surface forms which share only one feature and label this form pair with the shared (same) feature. Some example data is given in Table~\ref{tab:sharedFeat}. Since features depend on the language, the number of labels and the statistics of the dataset differ per language. \rev{}{The ability to detect shared morphological features is expected to contribute to the encoding of agreement.}
\begin{table}[!ht]
\center
      \begin{tabular}{l|p{4cm}|p{4cm}|c}  
      \textbf{L} & \textbf{form1} & \textbf{form2} & \textbf{SameFeat} \\
      \hline 
      TR & yalvaracaksınız \newline \small \textit{beg} \sc{(V.2Pl.Fut)} & onadı \newline \small \textit{approve} \sc{(V.3Sg.Pst)} & Polarity \\ \hline
      TR & yolculuklarına  \newline \small \textit{travel} \sc{(N.Poss3Pl.\textbf{Dat})} & düşmanıma \newline \small \textit{enemy} \sc{(N.Poss1Sg.\textbf{Dat})} & Case \\ \hline
      TR & taşımam \newline \small \textit{\textbf{carry}} \sc{(V.1Sg.Prs.Neg)} & taşıdılar \newline \small \textit{\textbf{carry}} \sc{(V.3Pl.Pst)} & Lemma \\ \hline
      TR & sarımsaklarım \newline \small \textit{garlic} \sc{(N.Pl.\textbf{Poss1Sg}.Nom)} & cümlemde \newline \small \textit{sentence} \sc{(N.Sg.\textbf{Poss1Sg}.Loc)} & Possession \\ \hline
      RU & pantera \newline \small \textit{panther} \sc{(N.Nom.\textbf{Sg})} & optimisticheskogo \newline \small \textit{optimistic} \sc{(Adj.Gen.\textbf{Sg})} & Number \\ \hline
      DE & Stofftiere \newline \small \textit{stuffed\_animal} \sc{(N.\textbf{Nom}.Pl)} & Tennisplatz \newline \small \textit{tennis\_court} \sc{(N.\textbf{Nom}.Sg)} & Case \\
      \end{tabular}
  \caption{Examples of form pairs with \textit{only} one shared feature. \emph{Poss3Pl}: possession by 3rd plural person, \emph{Poss1Sg}: possession by 1st singular person. Shared features shown in \textbf{bold}. Turkish positive polarity is not explicitly tagged by Unimorph. \emph{TR}: Turkish, \emph{RU}: Russian, \emph{DE}: German}
  \label{tab:sharedFeat}
\end{table}

\paragraph{OddFeat}
This test is the opposite of the shared feature test. We prepare pairs of surface forms which differ only by one feature value and label them with this odd feature. Some examples are given in Table~\ref{tab:oddFeat}. 
\begin{table}[!ht]
\center
      \begin{tabular}{l|p{4cm}|p{4cm}|c}  
      \textbf{L} & \textbf{form1} & \textbf{form2} & \textbf{OddFeat} \\
      \hline 
      TR & istemeyecek \newline \small \textit{want} \sc{(V.3Sg.Fut.\textbf{NEG})} & isteyecek \newline \small \textit{want} \sc{(V.3Sg.Fut)} & Polarity \\ \hline
      TR & seçenekler \newline \small \textit{option} \sc{(N.\textbf{Nom}.Pl)} & seçeneklere \newline \small \textit{option} \sc{(N.\textbf{Dat}.Pl)} & Case \\ \hline
      TR & iyileşiyorlardı \newline \small \textit{\textbf{heal}} \sc{(V.3Pl.Pst.Prog)} & geziyorlardı \newline \small \textit{\textbf{travel}} \sc{(V.3Pl.Pst.Prog)} & Lemma \\ \hline
      TR & deneyimlerine \newline \small \textit{experience} \sc{(N.Dat.Pl.\textbf{Poss2Sg})} & deneyimlerime \newline \small \textit{experience} \sc{(N.Dat.Pl.\textbf{Poss1Sg})} & Possession \\ \hline
      RU & zashitu \newline \small \textit{defence} \sc{(N.\textbf{Acc}.Sg)} & zashite \newline \small \textit{defence} \sc{(N.\textbf{Dat}.Sg)} & Case \\ \hline
      ES & legalisada \newline \small \textit{legalized} \sc{(V.Sg.Ptcp.\textbf{F})} & legalisado \newline \small \textit{legalized} \sc{(V.Sg.Ptcp.\textbf{M})} & Gender \\ \hline
      DE & integriert \newline \small \textit{\textbf{integrate}} \sc{(V.3sg.Ind.Prs)} & rechnet \newline \small \textit{\textbf{count}} \sc{(V.3Sg.Ind.Prs)} & Lemma \\
      \end{tabular}
  \caption{Examples of form pairs with \textit{only} one different feature. Odd features shown in \textbf{bold}. Turkish positive polarity is not tagged by Unimorph. \emph{Poss2sg}: possession by 2nd singular person, \emph{Poss1Sg}: possession by 1st singular person. Odd features shown in \textbf{bold}. Turkish positive polarity is not explicitly tagged by Unimorph. \emph{TR}: Turkish, \emph{RU}: Russian, \emph{ES}: Spanish, \emph{DE}: German}
  \label{tab:oddFeat}
\end{table}
\rev{}{Although these contrastive features are not directly linked to any simple linguistic property, we hypothesize that they can be valuable assets to compare/diagnose models for which it is important to learn the commonalities/differences between a pair of tokens, such as question answering, or natural language inference tasks.}



\subsection{Dataset Creation}
\label{ssect:testcreation}
 
\rev{}{In this section, we introduce the methodology for creating the type-level probing tasks, i.e., tasks where surface forms are probed without the context. Afterwards, the creation process for token-level probing tasks (i.e., where surface forms to be probed are provided within a context) is described. The focus of our study is on type-level tasks, however we provide a set of similar token-level tasks for comparison and discussion of future work.}

\subsection{Type-Level Probing Tasks}
\label{ssec:typeleveltasks}

     \rev{One of the goals of this study is to provide downstream performance prediction for as many languages as possible. Hence, w}{W}hile searching for a dataset to source the probing tests from, the number of languages this dataset covers is of key importance. Although there is only a small number of annotated \emph{truly} multilingual datasets such as Universal Dependencies, unlabeled datasets are more abundant such as Wikipedia~\footnote{https://www.wikipedia.org/} and Wiktionary~\footnote{https://www.wiktionary.org/}. For type-level probing tasks, we use \rev{Unimorph}{UniMorph~2.0~\cite{kirov-etal-2018-unimorph}} that provides a dataset of inflection paradigms with universal morphology features mapped from Wiktionary for many of the world's languages. In addition to UniMorph, we use the lexicon and the software provided by Wuggy to generate pseudowords~\cite{keuleers2010wuggy,erten2014turkish}. Finally we use word frequency lists extracted from Wikipedia. We follow different procedures to create datasets for each test type. Here, we briefly explain the creation process of single form feature tests such as \textit{Tense}, \textit{Voice}, \textit{Mood}; paired form feature tests: \textit{OddFeat} and \textit{SameFeat}; followed by \textit{Character Bin}, and pseudoword generation via Wuggy.  

    \paragraph{Single Form Feature Tests}
    
    A word annotated with UniMorph features can be used in several probing tests. For instance, the Turkish word ``grubumuzdan'', (\textit{from our groups}) is marked with the N.Sg.Poss1Pl.Abl tag and can be used to probe the \textit{POS}, \textit{Case marking}, \textit{Number} and the \textit{Possession} features since it has the N (Noun), Abl (Ablative), Sg (Singular) and Poss1Pl (Possession by first person plural) tags. While generating the tests, we check if the following conditions for a language and target feature are satisfied:
    \begin{itemize}
    \item Since we need to train classifiers for the probing tests, we need large enough training data. We eliminate the language/feature pair if total number of samples for that certain feature is less than 10K~\footnote{In our preliminary experiments, we found that 10K is large enough to provide sufficient clues for the linguistics classifier to predict linguistic labels; and small enough to cover as many languages as possible}.
    \item If a feature, e.g. \textit{case marker}, does not have more than one unique value for a given language-feature pair, it is excluded from the tests.
    \end{itemize}
    
    
    In addition, we perform two additional preprocessing steps: (1) removal of ambiguous forms with respect to linguistic feature, (2) partial filtering of the infrequent words. Ambiguity is one of the core properties of the natural language, and a single word form can have multiple morphological interpretations. For instance the German lemma ``Teilnehmerin'' would be inflected as ``Teilnehmerinnen'' as a plural noun marked either with accusative, dative or a genitive case marker. We remove such words with multiple interpretations for the same feature. \rev{This is a deliberate design choice we make, which, while potentially causing some systematic removals, substantially simplifies the task architecture and guarantees fair testing. In Fig.\ref{fig:amb}, we show the ratio of removed instances to all instances separately for each language and test. As can be seen, the highest ratio is for German and around 30\%, followed by the \emph{Gender} and the \emph{Case} tests for Russian.}{This is a deliberate design choice we make, which, while potentially causing some systematic removals for certain tasks such as German case, substantially simplifies the task architecture and guarantees fair testing. The ambiguity ratios are discussed in more details in Sec.~\ref{ssec:disc_prob}.}
    
    UniMorph dataset contains many grammatically correct but infrequent word forms such as the English ``transglycosylating'' or the Turkish ``satrançlarımızda'' (\textit{in our chesses}). To make sure that our probing tests are representative of the language use, we utilize the frequent word statistics extracted from the Wikipedia dump of the corresponding language. For each probing test, the dataset is compiled so that 80\% of the forms are frequently encountered words. We keep a portion of ``rare'' words~\footnote{We have followed the 80\% frequent versus 20\% rare word ratio. In some exceptional cases where we can not choose 80\% of the words from frequency dictionary due to small wikipedia, we allow a larger portion of rare words to have at least 10K instances.} and use a considerably large proportion of frequency dictionary, (e.g., we keep the first 1M words for Russian) to identify frequent words in order to keep our tests domain-independent, hence not provide any unjust advantage to embedding models trained on Wikipedia. Finally, \rev{to have a dataset with feasible baseline scores,}{} we introduce surface forms \rev{}{of ``None'' class, i.e., forms that} do not contain that test feature. For instance if the \textit{Tense} feature is probed, the 30\% of the probing dataset contains nominal forms that are from ``None'' class. \rev{}{Most NLP downstream tasks need to distinguish between a ``None'' class and other class labels. For instance an SRL model needs to decide whether a token is an argument of a predicate; a dependency parser needs to decide if two tokens can be connected by a dependency relation; or a NER needs to predict if a token is part of a named entity or not. We believe this setting provides a more realistic probing task scenario compared to having only the positive examples of a given linguistic feature.}
    
    
    \paragraph{Paired Feature Tests}
    Unlike for single features, we did not remove ambiguous forms for paired features tests, i.e., \textit{OddFeat} and \textit{SameFeat} due to the retrictive nature of the tests. For instance, while probing for the \textit{OddFeat} between two forms, we assume that there exists a word pair differing only by one feature. Therefore, here we only consider one certain interpretation of the word form, which would share $n-1$ features with the interpretation of the other form, where $n$ is the total number of UniMorph features in both words. 
    
    Dataset for this test type is created in two separate steps: (1)~for unimorph tags (2)~for lemmas. For the \textit{SameFeat}, we first group the words that contain the feature of interest together for the step (1). Then we split each feature group into two and sample $k=500$ words from both groups. These word pairs are compared against each other, and included in the test set if they share the same value for the feature of interest, but differ in all other features. Since some features are tagged by default, e.g., POS, we exclude these features from the comparison process. Otherwise our dataset would have no instances, since, for example, all nouns share the ``N'' tag. In addition to POS tags, we exclude the \textit{Mood} feature from Finnish and Turkish, and \textit{Interrogativity} feature from Turkish, since all verbs in UniMorph data share the same tag for those features. For (2), we follow the same steps, but check if the lemma values are the same and others are different.~\footnote{We perform similar preprocessing and dataset balancing for all languages. The details of parameter values can be found on the project website.}
    
    While preparing the dataset for the \textit{OddFeat}, we first group the words by \emph{lemma} tagged with the target feature for the step (1). Then we randomly sample elements from each lemma group, and perform pairwise comparison. If two sampled forms have different values for the feature (e.g., Ablative and Locative) but have the same set of values for the other features (e.g., Singular), then they are assigned this feature as the label. In addition to the features with different values, we also consider the features that are not explicitly tagged. For instance if only one of the forms has the \textit{Possession} feature, but all features except \textit{Possession} are shared among these two forms, we create a test pair with the value \textit{Possession}. To generate the test pairs for the step (2), we group the words by their feature sets, i.e., different forms with the exact same set of feature values will be clustered together. Then we split each group into two, and sample $k=100$ number of forms from both halves. 
    The procedure described above results in unbalanced datasets, usually dominated by the \textit{Number} feature. In order to avoid this, we sample proportionally from such overly sized feature test pairs. 
    
    \paragraph{Character Bin} 
    After removing the ambiguous forms, we have created bins of numbers for character counts since the variation was high. We used the following bins for character counts: [0-4, 5-8, 9-12, 13-16, 17-20, >20]. We applied the same bins for all languages. 
    
    \paragraph{Pseudo Word Test}
    Finally, we have generated pseudo words for 9 languages. To do so, we first sampled 10K in-vocabulary words from the lexical resources provided by Wuggy. We then use those words as seeds for the Wuggy generator, and generate pseudowords by setting the maximum number of candidates per word to 5, maximal search time per word to 5 seconds; and restricting the output to match the length of sub-syllabic segments, match the letter length, match transition frequencies and match 2/3 of sub-syllabic segments. 
    
    The sets of languages for each probing test introduced in Section~\ref{ssect:tesdefs} are given in Table~\ref{tab:tests_langs}. In total, we have created 15 probing tests for 24 languages, each containing 7K training, 2K development and 1K test instances.

\subsection{\rev{}{Token-Level Probing Tasks}}
\label{ssec:tokenleveltasks}
    
    \rev{}{Type-level probing has several advantages: it's compact and less prone to majority and domain shift effects. However, since downstream NLP tasks mostly operate on full-text data, decoupling evaluation from running text might result in a less realistic performance estimates; besides, it limits the evaluation of contextualized word representations and black-box models. To investigate the limitations and the strengths of type-level tasks, we prepare a set of comparable token-level probing tasks using the modified Universal Dependency Treebanks where the MSDs have been converted to the UniMorph schema~\cite{McCarthySCHY18}. Contrary to the type-level tasks, we do not filter out any infrequent or ambiguous surface forms; and we do not introduce a ``None" class for convenience. Since the dataset is annotated with the same schema as in our type-level tasks, we simply adapt our existing source code that creates single form feature tests (e.g., \emph{Tense}, \emph{Case}) for token-sentence pairs. Similar to the single form type-level tasks, if total number of samples for a certain feature is less than 10K; or if a feature, e.g. \emph{case marker} has only one value, we exclude that feature-language pair from the tests. The created tests have the sentence, word index and feature label information. As an example, the following line is taken from the Person-English test-language pair: \emph{``Looks good"}, \emph{0}, \emph{Third person Singular}; meaning that the word at index 0 in the given sentence (\emph{``Looks"}) has the \textsc{Third person Singular} label.} 
    
    \rev{}{Following the Sec.~\ref{ssec:typeleveltasks}, we have created the same single form feature tests for all available languages with the modified UD treebank; each containing 7K training, 2K development and 1K test instances. The current version of the token-based suite only contains category-based, morphological tests, however, it can be easily extended for other probing tasks: OddFeat, SameFeat, TagCount, and CharacterBin.}
    
\subsection{\rev{}{Discussion on probing task types}}
\label{ssec:disc_prob}
    \rev{}{Properties and quality of the token-level probing tasks are strongly tied to the properties and the quality of resources used while creating them. To provide more insights, Table~\ref{tab:uni_versus_ud} provides essential statistical information for type- and token-level task sets, focusing on the languages we experiment with (explained in Sec.~\ref{ssec:langchoice}) later in this work. 
    \begin{table}[!ht]
    \centering
    \scalebox{0.8}{
        \begin{tabular}{l@{\hskip 0.2in}cccc@{\hskip 0.5in}ccccc}
         & \multicolumn{4}{c}{Type-level} & \multicolumn{5}{c}{Token-level} \\
         \cmidrule{2-10}
          & $|\pi|$ & \#form & types & amb\% & \#sent & \#token & $|sent|$ & V(\%) & amb\% \\
        \midrule
        Finnish & 57K & 2.5M & N, V, A & \textbf{4.87} & 31K & 339K & 10.81 & 83K \textit{(4.07)} & \textbf{17.62} \\
        Turkish & 3.5K & 275K & N, V, A & \textbf{7.76} & 6K & 67K & 11.25 & 22K \textit{(3.06)} & \textbf{19.28} \\
        Russian & 28K & 474K & N, V, A & \textbf{12.51} & 63K & 1.1M & 17.89 & 135K \textit{(8.29)} & \textbf{23.75} \\
        German & 15K & 179K & N,V & \textbf{25.92} & 14K & 263K & 18.74 & 49K \textit{(5.39)}	& \textbf{27.47} \\
        Spanish & 5.5K & 383K & V & \textbf{10.75} & 30K & 883K & 29.12 & 68K \textit{(13.04)} & \textbf{35.1}\\
        \midrule
        \end{tabular}
        }
    \caption{\rev{}{Statistics for the resources used during the creation of type and token-level probing tasks. $|\pi|$: Number of inflection paradigms, \#form: Number of inflected forms, N: Noun, V: Verb, A: Adjective, amb\%: Ratio of ambiguous forms, \#sent: Number of sentences, \#token: Number of tokens, $|sent|$: Average sentence length, V (\%): Vocabulary size (\#token/V)}}
    \label{tab:uni_versus_ud}
    \end{table}}
    
    \rev{}{\paragraph{Dataset size} The agglutinative languages, Finnish and Turkish, have higher amount of instances for type-level tasks than token-level tasks. This is due to their productive morphology that enables generating large amount of surface forms from a single lemma. One can observe that for all fusional languages, number of tokens in token-level resource exceeds the number of forms available from type-level resource, while agglutinative languages follow an opposite trend. 
    This is due to practical reasons: Unimorph is based on Wiktionary data, and due to their agglutinative nature Finnish and Turkish allow easier generation of word forms to populate the paradigms, while fusional languages require manual annotation. At the same time, Russian, German and Spanish are higher-resourced langauges that offer large-scale treebanks from which the Universal Dependency Treebank data has been sourced.} \rev{}{\paragraph{Data Domain and Token Frequency} Type-level tasks are induced from a dictionary based resource (Wiktionary), while token-level tasks are based on existing language-specific treebanks. For instance the largest Turkish treebank (UD-IMST) is collected from daily news reports and novels; while the biggest Finnish treebank is a collection of manually annotated grammatical examples. Token-level tasks based on running text -- especially given that treebanks are often based on a homogeneous document collection -- are inevitably biased to the domain of this text, while type-level probing tasks are expected to be domain-neutral. In particular, dictionary-based tasks do not contain any frequency information of the surface forms, while token-level tasks do. Although the frequency information may be helpful in some cases, e.g., when the domains of the downstream tasks and the probing tasks are similar, it would also add a bias regarding the distribution of specific features. A token-level test would be penalized less for misclassifying rare forms, and the probing classifier might benefit from using majority class information which might depend on the domain \emph(e.g., singular nouns are more frequently observed than plural nouns).} \rev{}{\paragraph{Data quality} Wiktionary is a collaborative effort, while the UD Treebanks are mostly annotated by a handful of experts. Although we cannot find an exact measure on the accuracy of Wiktionary data, a dataset with large number of collaborators may have less annotation artifacts than a dataset created by a few experts. On the other hand, both datasets may have been effected negatively from an automatic conversion process, mostly due to converting language-specific features to universal tags.} \rev{}{\paragraph{Lexical variety} While Turkish, Finnish and Russian type-level UniMorph data have all lexical classes, German data does not include any adjectives. Furthermore, Spanish only has verb inflections that limit the scope of probing. On the other hand, treebanks for the token-level tests are based on running text in which all lexical classes are represented.} \rev{}{\paragraph{Ambiguity} In Table~\ref{tab:uni_versus_ud} we list the average number of ambiguous forms, i.e., forms that might be expressing more than one morphological feature bundle, for both task types. For type-level, we have averaged the ambiguity ratios over all probing tasks. For token-level tasks, we simply calculated the ratio of surface forms with different morphological tag sets to all surface forms. We notice that for agglutinative languages, where we have one-to-one morpheme to meaning mapping, the ambiguity ratios for both type- and token-level tests are lower. 
    For fusional languages the ambiguity ratios are higher, mostly due to syncretism - one word form might encode several morphological feature bundles. In particular for German, the average is around 26\% (before removal of ambiguous entries), which mean loss of considerable amount of data points. In general, token-level tests have higher ambiguity ratios, however since the tokens are provided within the context, it enables models to resolve the ambiguity.} \rev{}{\paragraph{Use Cases} Despite their similarities, type- and token-level probing tasks differ in terms of potential use cases and limitations. From the representation perspective, type-level tasks are better suited for probing context-free word embeddings (i.e., static or subword-level); while token-level tasks are more suitable for contextual embeddings (due to having many duplicate training and test instances when tokens are isolated from context). Token-level tasks can be used as a diagnostic probing tool for any downstream model layer that doesn't require any additional task-specific inputs (e.g. part-of-speech tags for dependency parsers, predicate flags for SRL). Type-level tasks, on the other hand, are more suited to diagnose the initial word encoding layer that generate a type of word representation in isolation; not the intermediate hidden layers that require contextual information.} 
    \rev{}{\paragraph{Summary} In summary, both token- and type-level probing test designs come with certain implications, and the choice of the probing set depends on the task at hand. Type-level probing tasks have the advantages of containing less bias (domain, annotator and majority class); while token-level tests might be sensitive to the domain biases from the underlying full-text data. On the other hand, token-level tests have the advantage of being more lexically diverse; while type-level tasks can be less diverse for some languages like Spanish, French and English. In terms of dataset sizes and the number of languages that can be covered, both type- and token-level probing tests are similar. Finally, type-level tasks are better suited to probe traditional word embeddings or initial word encoding layers that do not require contextual information; while token-level tasks are more suitable for probing contextual embeddings or intermediate layers if the layers do not require additional linguistic information.} 

\section{Evaluation Methodology}
In this section, we discuss our probing task evaluation methodology. First of all, due to the large number of languages and embedding models available, we choose a subset of each. We describe how we decide on the languages to evaluate on in Sec.~\ref{ssec:langchoice}\rev{, and the set of pretrained embeddings we have used is detailed in Sec.~\ref{ssec:embeds}.}{.} 
Next, in order to investigate the relation between probing and downstream tasks, we evaluate \rev{the \emph{same} set of embeddings}{a set of diverse multilingual embedding models} intrinsically via our probing tasks as explained in Sec.~\ref{ssec:intrinsic} and extrinsically on several downstream tasks discussed in Sec.~\ref{ssec:extrinsic}, and investigate the correlations between the corresponding task performances.
Finally, in Sec.~\ref{ssec:diagtool} we show how the proposed probing tests can be used as a diagnostic tool for black box NLP systems in a case study. 

\subsection{Languages}
\label{ssec:langchoice}

	We have identified a list of languages to test our hypotheses on various research questions such as the relation between downstream and probing tasks or the information encoded in layers of black box models. For this we have considered the following criteria: 
	\begin{itemize}
	\item Chosen languages should have relatively broad resource coverage \textit{e.g., annotated data for a variety of downstream tasks},  
	\item The set of chosen languages should have a high coverage of probing tests; and the number chosen languages should be in proportion to the number of languages that are probed for a certain test,
	\item The languages should be as typologically diverse as possible in terms of linguistic properties we are probing for. 
	\end{itemize}  
	Considering the above, for in-depth experimentation we have selected 5 languages -- German, Finnish, Turkish, Spanish and Russian, which are shown in colors in Table~\ref{tab:tests_langs}. Most of them have annotated resources in addition to Universal Dependencies treebanks, e.g., datasets created for named entity recognition (NER), \rev{}{Natural Language Inference (NLI)} and semantic role labeling (SRL). As can be seen from Table~\ref{tab:tests_langs}, all probing tests are covered and their ratio to other languages is well proportioned for each test. Our selected languages belong to diverse language families, namely from Germanic, Uralic, Turkic, Romance and Slavic\rev{.}{; and are typologically diverse, i.e., have representatives from agglutinative (Finnish and Turkish) and fusional (German, Spanish, Russian) languages.}  

	\begin{table} 
	  \caption{List of languages for each probing task. Languages shown in colored cells are the languages we experiment on. General refers to POS, Tag Count and Character Bin. Some of the tests with fewer number of languages are concatenated vertically for convenience.}
	  \label{tab:tests_langs}
	  \scalebox{0.6}{
		\begin{tabular}{*{9}{l}}
			\hline
			\cellcolor{gray!25}\textbf{CASE} & \cellcolor{gray!25}\textbf{MOOD}  & \cellcolor{gray!25}\textbf{NUMBER}  & \cellcolor{gray!25}\textbf{General} & \cellcolor{gray!25}\textbf{PERSON} & \cellcolor{gray!25}\textbf{POLARITY} & \cellcolor{gray!25}\textbf{TENSE} & \cellcolor{gray!25}\textbf{ODD FEAT} & \cellcolor{gray!25}\textbf{SAME FEAT} \\ 
			\hline
			arabic & arabic & armenian & arabic & arabic & portuguese & armenian & armenian & arabic \\ 
			armenian & armenian & catalan & armenian & armenian & \cellcolor{blue!25}turkish & bulgarian & czech & armenian \\\cline{6-6}
			bulgarian & catalan & \cellcolor{yellow!25}finnish & bulgarian & catalan & \cellcolor{gray!25}\textbf{POSSESSION} & catalan & \cellcolor{yellow!25}finnish & bulgarian \\\cline{6-6}
			czech & \cellcolor{yellow!25}finnish & french & catalan & \cellcolor{yellow!25}finnish & armenian & \cellcolor{yellow!25}finnish & \cellcolor{orange!25}german & catalan \\ 
			estonian & french & \cellcolor{orange!25}german & czech & french & quechua & french & hungarian & czech \\ 
			\cellcolor{yellow!25}finnish & \cellcolor{orange!25}german & hungarian & danish & \cellcolor{orange!25}german & \cellcolor{blue!25}turkish & \cellcolor{orange!25}german & macedonian & danish \\\cline{6-6} 
			\cellcolor{orange!25}german & hungarian & italian & estonian & hungarian & \cellcolor{gray!25}\textbf{VOICE} & hungarian & greek & dutch \\\cline{6-6} 
			hungarian & italian & macedonian & \cellcolor{yellow!25}finnish & italian & arabic & italian & polish & estonian \\ 
			macedonian & polish & polish & french & macedonian & bulgarian & macedonian & portuguese & \cellcolor{yellow!25}finnish \\ 
			greek & portuguese & portuguese & \cellcolor{orange!25}german & greek & \cellcolor{yellow!25}finnish & greek & quechua & french \\ 
			polish & romanian & \cellcolor{red!25}russian & hungarian & polish & \cellcolor{red!25}russian & polish & romanian & \cellcolor{orange!25}german  \\ 
			quechua & serbian & \cellcolor{green!25}spanish & italian & portuguese & serbian & portuguese & \cellcolor{red!25}russian & italian \\ 
			\cellcolor{red!25}russian & \cellcolor{green!25}spanish & swedish & macedonian & quechua & swedish & quechua & serbian & macedonian \\\cline{6-6}\cline{3-3} 
			serbian &  & \cellcolor{gray!25}\textbf{GENDER}  & greek & romanian & \cellcolor{gray!25}\textbf{PSEUDO} & romanian & \cellcolor{green!25}spanish & greek\\\cline{6-6}\cline{3-3} 
			swedish &  & arabic & polish & \cellcolor{red!25}russian & basque & \cellcolor{red!25}russian & swedish & polish \\ 
			\cellcolor{blue!25}turkish &  & bulgarian & portuguese & serbian & dutch & serbian & \cellcolor{blue!25}turkish & portuguese \\ 
			 &  & macedonian & quechua & \cellcolor{green!25}spanish & english & \cellcolor{green!25}spanish &  & quechua \\ 
			 &  & greek & romanian & \cellcolor{blue!25}turkish & french & \cellcolor{blue!25}turkish &  & romanian \\ 
			 &  & polish & \cellcolor{red!25}russian &  & \cellcolor{orange!25}german  &  &  & \cellcolor{red!25}russian \\ 
			 &  & portuguese & serbian &  & serbian  &  &  & serbian \\ 
			 &  & \cellcolor{red!25}russian & \cellcolor{green!25}spanish &  & \cellcolor{green!25}spanish&  &  & \cellcolor{green!25}spanish \\ 
			 &  & serbo & swedish &  & \cellcolor{blue!25}turkish  &  &  & swedish \\ 
			 &  & \cellcolor{green!25}spanish & \cellcolor{blue!25}turkish &  & vietnamese &  &  & \cellcolor{blue!25}turkish 
		\end{tabular}
		}
	\end{table}

\subsection{Multilingual Embeddings}
\label{ssec:embeds}
	\rev{}{Following \namecite{aggarwal2016common} that discussed the need for having \textbf{diverse} and \textbf{heterogeneous} samples to conduct a correlation study, we have picked the multilingual embeddings that are trained with different objectives, architectures and units; and avoided using similar models trained with a slightly different hyperparameter (e.g., word2vec trained with the same settings except dimensionality) to avoid having subclusters in our samples.} \rev{We choose the most commonly used pretrained word embedding models that are available for a large number of languages, since they are the ones that will most likely to be used by a researcher working on multilingual problems.}{} Namely, in this work we experiment with the following word embedding models: word2vec \citep{Mikolov13}; fastText \citep{bojanowski:TACL2017}; GloVe with Byte Pair Encoding \citep[\rev{BPE}{GloVe-BPE} ;][]{sennrich:ACL2016}; supervised MUSE \citep{conneau2017word}; and ELMo \citep{peters:NAACL2018}. 

	\textit{word2vec} Among the selected representations, only \textit{word2vec} uses word as the basic unit. We have traind a word2vec model for each of the selected languages on the latest preprocessed (tokenized, lowercased) Wikipedia dump using 300-dimensional CBOW, a window of size 10 and minimum target count as 5. We have used the implementation provided by the authors~\footnote{\url{https://code.google.com/archive/p/word2vec/}}.  

	\textit{fastText} provides word representations that have subword-level information learned from character n-grams. In simple terms, words are represented as a linear combination of the character n-gram embeddings of the token's character n-grams. We use the embeddings distributed by fastText~\footnote{\url{https://fasttext.cc/}} which are trained on preprocessed Wikipedia using CBOW with position-weights, in dimension 300, with character n-grams of length 5, a window of size 5. 
	
	\rev{\textit{BPE}}{\textit{GloVe-BPE}} is another type of subword-level embedding that uses unsupervised morphological segments generated by a compression algorithm inspired from~\citet{Gage:1994}. We use the pretrained embeddings by \citet{heinzerling2018bpemb} which are trained on preprocessed Wikipedia using GloVe~\cite{glove:14}. We use the python wrapper open sourced by the authors~\footnote{\url{https://github.com/bheinzerling/bpemb}} with default dictionary size \rev{}{of 10K} and dimension 300. Since the tool provides embeddings for each segment, in case of multiple segments per token, we used the averaged vector as the word representation. 

	\textit{MUSE-supervised} embeddings are \emph{crosslingual} fasttext embeddings. \rev{which are actually monolingual fasttext embeddings aligned in a common space using ground-truth bilingual dictionaries.}{These embeddings are generated by aligning the monolingual fasttext embeddings in a common space (in our case English) using ground-truth bilingual dictionaries.} We used \rev{the pre-aligned}{the aligned and mapped} vectors distributed by the authors~\footnote{\url{https://github.com/facebookresearch/MUSE}}. The \rev{monolingual}{crosslingual} embeddings have the same \rev{}{technical} properties as the \textit{fastText} vectors described above. Since the authors only release the static embedding vector without the model, we could not generate embeddings for OOV words.  

	\textit{ELMo} embeddings are computed on top of two-layer bidirectional language models which use characters composed using convolutional neural networks (CNN). Unlike previously introduced embedding models, ELMo provides \textit{contextualized} embeddings, i.e., the same words would have different representations when used in different contexts. However, our probing tests are type-level (as opposed to token-level), thus we only use the representations generated independently per each token both for the intrinsic and extrinsic experiments. In scope of this study, ELMo embeddings are treated as powerful pretrained character-level \textit{decontextualized} vectors. To highlight this important detail, we further refer to our ELMo-derived embeddings as \emph{Decontextualized ELMo (D-ELMo)}. We use the multilingual pretrained ELMo embeddings distributed by the authors~\citep{che-EtAl:2018:K18-2,fares-EtAl:2017:NoDaLiDa}~\footnote{\url{https://github.com/HIT-SCIR/ELMoForManyLangs}}, which are trained with the same hyperparameter settings as the original \citep{peters:NAACL2018} for the bidirectional language model and the character CNN. They are trained on randomly sampled 20 million words from Wikipedia dump and Common Crawl datasets and have the dimensionality of 1024. We use the 3-layer averaged ELMo representation for each word. 

	For all the experiments described in Sec.~\ref{ssec:extrinsic} and Sec.~\ref{ssec:intrinsic}, we first created the vocabulary for all intrinsic and extrinsic datasets per language. Then, we generated the vectors using the embeddings that can handle OOV words, namely \emph{fasttext}, \rev{\emph{BPE}}{\emph{GloVe-BPE}} and \emph{D-ELMO}, for each language-intrinsic and language-extrinsic pair. The static embeddings: \emph{word2vec} and \emph{MUSE} are used as provided. Hence, for the models using these embeddings, each unknown (OOV) word is replaced by the UNK token and the same vector is used for all UNK words.   

\subsection{Intrinsic evaluation: probing tasks}
\label{ssec:intrinsic}

	Following \citet{senteval18}, we use diagnostic classifiers \citep{shi-etAl:ACL2016,adi2017fine} for our main probing tests. Our diagnostic classifier is a feedforward neural network with one hidden layer, followed by a ReLU non-linearity. The classifier takes as an input a fixed trained word vector and predicts a particular label specific to the probing test. For \textit{OddFeat} and \textit{SameFeat}, since the input consists of two words, we first concatenate both word vectors before feeding them into the feedforward network. For all tests, we use the same hyperparameters: 300 hidden dimension and 0.5 dropout rate. We train each model for 20 epochs with early stopping (patience=5). The input dimension vector depends on the type of pre-trained word vectors that will be evaluated. Our evaluation suite is implemented using the AllenNLP library \citep{Gardner2017AllenNLP}.

\subsection{Extrinsic evaluation: downstream tasks}
\label{ssec:extrinsic}
	We consider five tasks for our extrinsic evaluation: universal POS-tagging (POS), dependency parsing (DEP), named entity recognition (NER), semantic role labeling (SRL), and cross-lingual natural language inference (XNLI). The former two tasks are useful to measure correlation of our probing test sets to downstream syntactic tasks, while the latter three provide insight into the performance on more semantic tasks. Since our main goal is to evaluate the quality of the \textit{pre-trained} word embedding spaces, we neither update the word vectors during training nor use extra character-level information. Except for SRL, all tasks described below are trained using the models implemented in AllenNLP library.

	\paragraph{POS Tagging} This is a classic sequence tagging task, where the goal is to assign a sequence of POS tags given the input sentence. We use data from the Universal Dependencies (UD) project version 2.3 \citep{UD2.3}, and adopt universal POS tags as our target labels.  For the tagging model, we use a bidirectional LSTM encoder with 300 hidden units and 0.5 dropout. We use Adam optimizer with initial learning rate 0.001. We train each model with mini-batch size of 32 for 40 epochs, with early stopping (patience=10). We use the accuracy as our performance metric. \rev{}{It must be noted that the POS-tagging downstream task is different from the POS probing task: probing is a single-item, type-level classification task using a simple MLP classifier, while extrinsic POS is a sequence tagging task utilizing a more powerful Bi-LSTM architecture and operating on sentence level}.

	\paragraph{Dependency Parsing} The aim of dependency parsing is to predict syntactic dependencies between words in a sentence in the form of a tree structure. This task is especially interesting because of its deep interaction with morphology, which we will evaluate in our probing tests. We employ a deep biaffine parser of \citet{Dozat2016DeepBA}, which is a variant of graph-based dependency parser of \citet{mcdonald-crammer-pereira:2005:ACL}. The parsing model takes as input a sequence of token embeddings concatenated with the corresponding universal POS embeddings. The input is then processed by a multi-layer biLSTM. The output state of the final LSTM layer is then fed into four separate ReLU layers to produce four specific word representations: two for predicting the arcs (\textit{head predictions}) and another two for predicting the dependency label (\textit{label prediction}). The resulting four representations are used in two biaffine classifiers, one predicting the arc and another one to predict a dependency label, given a dependent/head word pair. For our experiments, we use 2 layer biLSTM with 250 hidden units, POS embedding dimension 100, and ReLU layer (for arc and label representations) with dimension 200. We train the model with mini-batch size of 128 for 30 epochs, and perform early stopping when the Label Attachment Score (LAS) on development set does not improve after 5 epochs.

	\paragraph{Named entity recognition} The goal of this task is to label the spans of input text with entity labels, e.g., \textit{Person}, \textit{Organization}, or \textit{Location}. Unlike POS tagging, NER annotates text spans and not individual tokens; this is usually represented via a (Begin, Inside, Outside) BIO-like encoding. We employ a standard NER architecture, a BiLSTM-CRF model where the output of BiLSTM is processed by a conditional random field to enforce global sequence-level constraints~\cite{huang2015bidirectional}. We use a 2-layer BiLSTM with 200 hidden units and 0.5 dropout trained for 20 epochs with patience 10, the performance is measured via span-based F1 score.



	\paragraph{Semantic Role Labeling} Semantic Role Labeling (SRL) is the automatic process of identifying predicate-argument structures and assigning meaningful labels to them. An SRL-annotated sentence with the predicate sense ``buy.01: purchase'' is shown below. 
	\begin{quote} 
	\small
	$[$Mark$]$\textsubscript{Arg0: Buyer} $[$bought$]$\textsubscript{buy.01} $[$a car$]$\textsubscript{Arg1: Thing bought} from $[$a retailer store$]$\textsubscript{Arg2: Seller}
	\end{quote} 
	We consider the dependency-based i.e., CoNLL-09 style, PropBank SRL, where the goal is to label semantic argument heads with semantic roles. We use the subword-level end-to-end biLSTM based sequence tagging SRL model introduced by \namecite{sahin:acl18}. It can either use pretrained embeddings as word representations, or learn task specific subword-level (character, character-ngram, morphology) representations by composing word vectors via a separate bi-LSTM network. Here, we only used pretrained word embeddings concatenated with a binary predicate flag (1 if the token is predicate, 0 otherwise) and 2 layers of bi-LSTMs with 200 hidden dimensions on top of these representations. Finally, tokens are assigned the most probable semantic role calculated via the final softmax layer. Weight parameters are initialized orthogonally, batch size is chosen as 32, and optimized with stochastic gradient descent with adaptive learning rate initialized as 1. Gradient clipping and early stopping with patience 3 are used. We use the standard data splits and evaluate the results with the official evaluation script provided by CoNLL-09 shared task. We report the role labeling F1 scores.

	\paragraph{Natural Language Inference} The NLI task aims to extract the relations such as Entailment, Neutral, and Contradiction between a pair of sentences -- a \emph{hypothesis} and a \emph{premise}. \rev{}{This objective has been formerly addressed in scope of the Recognizing Textual Entailment (RTE) task that used the resources provided by RTE challenge tasks which had a small size~\footnote{\url{https://aclweb.org/aclwiki/Textual_Entailment_Resource_Pool}}. Later a larger dataset, a.k.a. Stanford Natural Language Inference \citep[SNLI;][]{snli:emnlp2015} dataset, which has been compiled from English image caption corpora and labeled via crowdsourcing, has been introduced.} Some example pairs of sentences are shown in Table~\ref{tab:xnli_exs}. 
	\begin{table}[!ht]
	\center
	      \begin{tabular}{|p{5cm}|p{5cm}|c|} 
	      \hline 
	      \textbf{Premise} & \textbf{Hypothesis} & \textbf{Label} \\
	      \hline 
	      Met my first girlfriend that way. & I didn’t meet my first girlfriend until later. & Contradiction \\\hline
	      I am a lacto-vegetarian. & I enjoy eating cheese too much to abstain from dairy. & Neutral \\\hline
	      At 8:34, the Boston Center controller received a third transmission from American 11 & The Boston Center controller got a third transmission from American 11 & Entailment \\
	      \hline
	      \end{tabular}
	  \caption{Example sentence pairs taken from~\citep{N18-1101}}
	  \label{tab:xnli_exs}
	\end{table}
	As stated by ~\citet{snli:emnlp2015} and also can be seen from Table~\ref{tab:xnli_exs}, a high-performing NLI model should handle phenomena like tense, modality and negation, which are mostly covered by our probing tasks. 

	\rev{}{MultiGenre NLI \citep[MultiNLI;][]{N18-1101} is a recent dataset that covers a wider variety of text styles and topics}. The Cross-lingual NLI \citep[XNLI;][]{conneau2018xnli} dataset has been derived from MultiNLI and is used as a benchmark for evaluating cross-lingual sentence representations. This evaluation benchmark originally aimed at testing the models trained for the source language (English), on the target language, and covers 15 languages including Spanish, Turkish, Russian and German. It should be noted that the development and test splits for each language in XNLI have been translated by professional translators. The authors also release the automatic translation of MultiNLI training split which they use to align the cross-lingual sentence embeddings. Since the multilingual embeddings used in this study are not all cross-lingual, here we train a separate monolingual NLI model for each language by using the automatic translation data. We use the Enhanced LSTM model \citep[ESIM;][]{DBLP:conf/acl/ChenZLWJI17} with default parameters provided by AllenNLP framework. \rev{}{This model employs a sequential inference based on chain (bidirectional) LSTMs with attentional input encoding, enhanced with syntactic parsing information. In our experiments, we use pre-trained word embeddings to represent both hypothesis and premise tokens. These embeddings are kept fixed during training (not updated).}

\subsection{Diagnostic evaluation: a case study on SRL}
\label{ssec:diagtool}
	Another proposed application of our probing tests is to diagnose the layers of a black-box NLP model. In order to do so, we used the same SRL model as described in extrinsic evaluation (see Sec.~\ref{ssec:extrinsic}). This time, instead of using pretrained embeddings, we used randomly initialized character trigram embeddings. The model generates intermediate word representations by summing the weighted forward and backward hidden states from the character trigram bi-LSTM network. As the model is trained with a negative log likelihood loss for semantic roles, it is expected to learn character trigram embeddings and other model parameters that are better suited for SRL. In order to diagnose whether it \emph{does} indeed extract morphologically relevant information during training, we save the model states for different epochs and generate the word representations via the aforementioned internal biLSTM layer and use our intrinsic evaluation suite from Sec.~\ref{ssec:intrinsic}, to evaluate these representations. As preprocessing, all tokens are lowercased and marked with start and end characters. One layer of bi-LSTMs both for subword composition and argument labeling with hidden size of 200 are used. Character trigrams are randomly initialized as 200-dim vectors. The other hyperparameters are kept the same as Sec.~\ref{ssec:extrinsic}.

\section{Experiments and Results}

In this section, we first discuss the datasets used for our intrinsic and extrinsic experiments. We then provide the results and briefly discuss the general patterns and exceptions observed in both experiments. \rev{}{It is important to note that our primary goal is to compare the performance of embedding model \textit{instances}, and not of the embedding models per se: given that the performance of a particular trained instance might depend on a variety of factors such as dimensionality, preprocessing details and underlying textual corpora, a claim that a certain embedding method (e.g. word2vec) outperforms another embedding method in general would be far-fetched and would fall out of scope of our current study: instead we provide a toolkit that allows to empirically investigate the performance of the embedding spaces, which are by themselves treated as black box.} 

\subsection{Dataset}
\label{ssec:dataset}

    For intrinsic evaluation, we use the probing datasets that have been described in Sec~\ref{ssect:testcreation}, and experiment with the five languages: Finnish (Uralic), German (Germanic), Spanish (Romance), Russian (Slavic), and Turkish (Turkic) as discussed in Sec.\ref{ssec:langchoice}. 
    \rev{}{For POS tagging and dependency parsing, we use datasets from Universal Dependencies version 2.3 \citep{UD2.3}}. For the NER dataset, the Turkish and Russian data are substantially larger than the other languages. For practical reasons and fair comparison, we randomly sample 5-8\% subsets of the original datasets and split them into train/dev/test sets. \rev{}{The details of each UD treebank and other extrinsic dataset sources along with their statistics are presented in Table \ref{tab:extrinsic-data-stat}}.~\footnote{\rev{}{ Finnish NER data is available from \url{https://github.com/mpsilfve/finer-data} and the article ``A Finnish News Corpus for Named Entity Recognition'' where the dataset is described is reported to be under review.}}
    
    \begin{table}[!ht]
        \centering
        \scalebox{0.7}{
        \begin{tabular}{lllrrrrr}
        \multirow{2}{*}{Task} & \multirow{2}{*}{Language} & \multirow{2}{*}{Source} & \multicolumn{3}{c}{Number of tokens} & \multicolumn{2}{c}{OOV\%} \\
        \cmidrule{4-8}
        & & & train & dev & test & dev & test \\
        \midrule
        POS Tagging & Finnish & Finnish-TDT & 162.6K & 18.3K & 21.K	& 22.99 & 22.3 \\
        Dependency Parsing & German & German-GSD & 263.8K & 12.5K & 16.5K & 9.67 & 10.76 \\
        & Russian & Russian-SynTagRus & 870.5K & 118.5K & 117.3K & 8.44 & 8.68 \\
        & Spanish & Spanish-AnCora & 444.6K & 52.3K & 52.6K & 4.92 & 4.91 \\
        & Turkish & Turkish-IMST & 37.9K & 10.K & 10.K & 24.14 & 23.04 \\
        \midrule
        NER & German & Germeval-2014~\citep{benikova7germeval} & 452.9K & 41.7K & 96.5K & 11.34 & 11.29 \\
        & Russian & WikiNER \citep{I17-1042} & 169.1K & 55.4K & 55.2K & 16.65 & 16.75 \\
        & Turkish & TWNERTC \citep{Sahin2017AutomaticallyAT}  & 272.1K & 91.3K & 90.9K & 14.48 & 14.97 \\
        & Spanish & CoNLL-2002~\citep{sang02} & 264.7K & 52.9K & 51.5K & 7.43 & 5.63 \\
        & Finnish & FinNER & 180.1K & 13.6K & 46.4K & 18.9 & 19.7 \\
        \midrule
        SRL & Finnish & Finnish PropBank \citep{Haverinen2015} & 162.7K & 9.2K & 9.1K & 22.77 & 23.05 \\
        & German & CoNLL-09 \citep{conll09} & 648.7K & 32.K & 31.6K & 8.43 & 8.69 \\
        & Spanish & CoNLL-09 \citep{conll09} & 427.4K & 50.4K & 50.6K & 6.06 & 6.16 \\
        & Turkish & Turkish PropBank \citep{sahin17:lre} & 44K & 9.7K & 9.3K & 22.79 & 21.82 \\
        \midrule
        XNLI & German & \multirow{4}{*}{XNLI \citep{conneau2018xnli}} & 13.7M & 77.1K & 156.K & 5.46 & 5.57 \\
        & Russian & & 12.3M & 70.9K & 143.7K & 7.61 & 7.75 \\
        & Spanish & & 13.8M & 81.8K & 165.2K & 3.17 & 3.15 \\
        & Turkish & & 10.4M & 62.4K & 126.6K & 10.15 & 10.3 \\
        \midrule
        \end{tabular}
        }
        \caption{Sources and statistics of our extrinsic dataset. \rev{}{The NER datasets for Turkish and Russian are down-sampled.}}
        \label{tab:extrinsic-data-stat}
    \end{table}

\subsection{Results}
\label{ssec:results}
    \secrev{}{We first provide the results of intrinsic and extrinsic experiments for the type-level probing tasks. Later, we provide a comparison between token-level and type-level tests using the results of intrinsic experiments.}
    \subsubsection{\secrev{}{Results on Type-Level Probing Tasks}}
        We present the \secrev{}{type-level} probing test results of the multilingual embeddings introduced in Sec.~\ref{ssec:embeds} for each language/test pair in Table \ref{tab:all-probing-results}. In addition, we report the baseline scores calculated with majority voting baseline for each language/test pair. According to Table~\ref{tab:all-probing-results}, the majority of the tests had a baseline score under 50\%, although some language/test pairs had higher baselines due to the dataset properties such as lacking annotations for certain tags. These tests are \textit{POS} for Finnish, Spanish and Turkish and \textit{TagCount} for Finnish. In addition, \textit{SameFeat} and \textit{OddFeat} have relatively low baseline scores consistently across languages, generally followed by \textit{Case}. 

        Table~\ref{tab:all-probing-results} shows that all embedding models investigated in this work achieved their lowest score for \textit{CharacterBin}, \rev{followed by the \textit{OddFeat}}{}. \rev{}{As none of our embedding models use characters as basic units (except for D-ELMo which employs character-level CNN), it might be difficult for them to predict the number of characters from the surface form alone. However, we note that models that use subword units such as fastText, Glove-BPE, and D-ELMo in general obtain better performance than models with words as basic units (word2vec and MUSE).}
        \begin{table}[!ht]
            \centering
            \caption{Type-Level Probing task results for all languages. \textbf{Bold} represents the best score, while \textit{italics} is the second best.}
            \label{tab:all-probing-results}
            \scalebox{0.6}{
            \begin{tabular}{lcccccc}
            \hline
            \multicolumn{7}{c}{\textit{Finnish}} \\
            \hline
            \toprule
            Task & baseline & MUSE & word2vec & \rev{BPE}{GloVe-BPE} & fastText & D-ELMo \\
            \midrule
            Case & 30.0 & 49.3 & 59.9 & 83.4 & \textit{86.6} & \textbf{96.7} \\
            Mood & 50.0 & 62.3 & 67.9 & 84.7 & \textit{89.0} & \textbf{93.8} \\
            Number & 45.6 & 60.4 & 69.4 & 83.4 & \textit{90.3} & \textbf{97.4} \\
            POS & 67.9 & 75.3 & 70.3 & 85.7 & \textit{90.0} & \textbf{97.1} \\
            Person & 30.1 & 54.0 & 66.8 & 84.6 & \textit{88.8} & \textbf{94.6} \\
            Tense & 40.9 & 65.4 & 73.4 & 86.0 & \textit{90.6} & \textbf{94.7} \\
            Voice & 50.8 & 63.4 & 70.8 & 86.8 & \textit{89.6} & \textbf{95.1} \\
            \midrule
            CharacterBin & 44.2 & 45.0 & 44.8 & 52.0 & \textit{58.4} & \textbf{63.8} \\
            TagCount & 86.0 & 88.6 & 87.0 & 91.0 & \textit{95.0} & \textbf{98.4} \\
            \midrule
            OddFeat & 22.7 & 24.4 & 24.5 & 65.1 & \textit{76.7} & \textbf{88.4} \\
            SameFeat & 29.1 & 94.1 & 92.0 & \textit{96.9} & 96.5 & \textbf{98.4} \\
            \bottomrule
            \hline
            \multicolumn{7}{c}{\textit{German}} \\
            \hline
            \toprule
            Task & baseline & MUSE & word2vec & \rev{BPE}{GloVe-BPE} & fastText & D-ELMo \\
            \midrule
            Case & 34.2 & 62.0 & 68.7 & 90.9 & \textbf{95.1} & \textit{94.0} \\
            Mood & 37.4 & 54.3 & 54.1 & 90.1 & \textit{91.0} & \textbf{93.9} \\
            Number & 40.1 & 60.4 & 66.8 & 90.7 & \textit{93.7} & \textbf{97.7} \\
            POS & 55.8 & 63.1 & 65.8 & 92.2 & \textit{94.9} & \textbf{96.9} \\
            Person & 52.9 & 65.2 & 60.3 & 90.4 & \textit{91.5} & \textbf{95.8} \\
            Pseudo & 50.0 & 96.7 & 80.1 & 83.2 & \textit{90.0} & \textbf{91.0} \\
            Tense & 52.9 & 73.1 & 71.5 & 91.5 & \textit{92.9} & \textbf{93.2} \\
            \midrule
            CharacterBin & 45.4 & 49.0 & 45.0 & \textit{63.0} & 62.9 & \textbf{70.4} \\
            TagCount & 54.9 & 61.5 & 63.1 & 83.0 & \textit{86.5} & \textbf{89.2} \\
            \midrule
            OddFeat & 22.6 & 37.9 & 34.8 & 65.1 & \textit{71.2} & \textbf{75.4} \\
            SameFeat & 28.4 & 84.5 & 86.5 & \textit{89.6} & \textbf{90.4} & 89.0 \\
            \bottomrule
            \hline
            \multicolumn{7}{c}{\textit{Spanish}} \\
            \hline
            \toprule
            Task & baseline & MUSE & word2vec & \rev{BPE}{GloVe-BPE} & fastText & D-ELMo \\
            \midrule
            Gender & 34.5 & 67.0 & 74.5 & 98.0 & \textit{98.8} & \textbf{99.8} \\
            Mood & 52.0 & 67.0 & 66.1 & 89.2 & \textit{90.9} & \textbf{95.0} \\
            Number & 34.0 & 69.2 & 69.9 & \textit{95.0} & \textit{95.0} & \textbf{99.8} \\
            POS & 70.9 & 85.6 & 84.1 & 97.6 & \textit{98.5} & \textbf{99.6} \\
            Person & 27.4 & 60.9 & 52.8 & \textit{92.6} & 87.8 & \textbf{98.6} \\
            Pseudo & 49.8 & \textit{92.3} & 89.4 & 75.9 & 91.9 & \textbf{94.7} \\
            Tense & 39.9 & 59.1 & 60.8 & \textit{87.1} & 85.9 & \textbf{95.0} \\
            \midrule
            CharacterBin & 50.9 & 55.2 & 55.3 & \textit{72.3} & 69.6 & \textbf{76.2} \\
            TagCount & 40.0 & 61.0 & 59.0 & \textit{90.8} & 87.8 & \textbf{95.8} \\
            \midrule
            OddFeat & 44.8 & 53.4 & 55.8 & 77.1 & \textit{78.5} & \textbf{81.7} \\
            SameFeat & 27.2 & 89.6 & 89.1 & \textit{91.1} & \textbf{93.3} & \textit{91.1} \\
            \bottomrule
            \hline
            \multicolumn{7}{c}{\textit{Russian}} \\
            \hline
            \toprule
            Task & baseline & MUSE & word2vec & \rev{BPE}{GloVe-BPE} & fastText & D-ELMo \\
            \midrule
            Case & 31.0 & 57.3 & 78.0 & \textit{80.8} & 62.0 & \textbf{96.7} \\
            Gender & 39.8 & 57.7 & 78.3 & \textit{95.4} & 80.7 & \textbf{99.3} \\
            Number & 41.1 & 54.7 & 75.7 & \textit{89.7} & 74.3 & \textbf{96.9} \\
            POS & 48.4 & 56.5 & 67.8 & \textit{89.7} & 74.2 & \textbf{98.2} \\
            Person & 31.9 & 49.4 & 72.2 & \textit{93.0} & 81.0 & \textbf{96.7} \\
            Tense & 43.8 & 56.3 & 73.6 & \textit{90.1} & 73.6 & \textbf{94.3} \\
            Voice & 47.6 & 62.2 & 66.5 & \textbf{99.4} & 96.1 & \textit{99.0} \\
            \midrule
            CharacterBin & 46.0 & 46.3 & 52.5 & \textit{68.9} & 64.4 & \textbf{70.9} \\
            TagCount & 53.8 & 60.4 & 68.5 & \textit{85.2} & 67.9 & \textbf{96.4} \\
            \midrule
            OddFeat & 21.8 & 36.9 & 48.2 & \textit{74.4} & 55.4 & \textbf{90.0} \\
            SameFeat & 29.4 & 84.7 & 90.9 & \textit{93.9} & 93.6 & \textbf{97.6} \\
            \bottomrule
            \hline
            \multicolumn{7}{c}{\textit{Turkish}} \\
            \hline
            \toprule
            Task & baseline & MUSE & word2vec & \rev{BPE}{GloVe-BPE} & fastText & D-ELMo \\
            \midrule
            Case & 31.1 & 63.5 & 57.4 & \textit{87.1} & 85.4 & \textbf{96.1} \\
            POS & 75.5 & 85.9 & 83.5 & \textit{95.9} & 94.8 & \textbf{98.4} \\
            Person & 30.3 & 52.5 & 52.0 & \textit{93.5} & 90.5 & \textbf{96.1} \\
            Polarity & 44.6 & 62.0 & 61.0 & \textbf{97.3} & 93.6 & \textit{96.1} \\
            Possession & 30.6 & 59.2 & 56.7 & \textit{87.1} & 75.5 & \textbf{92.5} \\
            Pseudo & 51.5 & \textit{90.3} & 90.2 & 71.4 & 79.6 & \textbf{91.7} \\
            Tense & 34.9 & 57.7 & 58.8 & \textit{89.4} & 85.4 & \textbf{94.7} \\
            \midrule
            CharacterBin & 46.1 & 58.1 & 53.6 & \textit{66.7} & \textit{66.7} & \textbf{71.5} \\
            TagCount & 46.6 & 71.4 & 60.7 & \textit{85.6} & 79.9 & \textbf{89.8} \\
            \midrule
            OddFeat & 38.7 & 38.5 & 40.6 & 76.7 & \textbf{79.8} & \textit{79.0} \\
            SameFeat & 21.3 & 73.9 & 74.7 & \textit{86.9} & \textbf{90.0} & 86.5 \\
            \bottomrule
            \end{tabular}
            }
        \end{table} 
        In order to assess the difficulty of the tests, \rev{we calculated}{one can calculate} the gap between the average performance of the embeddings and the baseline scores. A small gap \rev{can point}{points} to a ``hard-to-beat'' majority vote baseline. \rev{It can also point to a probing test that would be difficult to \rev{master}{be learned by the word embedding models}}{} After eliminating the tests with high baseline scores, we observe that majority of the tests have seen improvements ranging between 50\%-200\%, albeit their low baseline scores. 


        First of all, for probing tests we observe that all embeddings outperform the baseline for all tasks and languages. Apart from a few cases, we see that \emph{D-ELMo} achieves the highest scores \rev{}{in probing} for all language-test pairs, generally followed by \emph{fastText} and \rev{\emph{BPE}}{\emph{GloVe-BPE}}. \rev{}{There are several factors that can explain why \emph{D-ELMo} achieves the highest scores. First of all, \emph{D-ELMO} models are trained on a different text source (subset of Wikipedia combined with CommonCrawl). Second, D-ELMo had additional training objectives compared to other traditional language modeling ones. Third, unlike other embedding models, D-ELMo is the only model with a layer operating on character-level. It has been shown several times in separate studies~\cite{sahin:acl18,vania2017} that character-level models perform better than other subword-level models on a number of downstream tasks. Fourth, it has dimensionality of 1024 while other models are of dimension 300. Finally, it may be a combination of all properties explained above. A careful investigation of the exact property of D-ELMo that grants it advantage falls outside of the scope of this study.} For the languages Finnish, Russian and Turkish, \emph{D-ELMo} outperforms the other embeddings by a larger margin compared to Spanish and German. \emph{fastText} and \emph{GloVe-BPE} perform similarly, except from Russian where \rev{\emph{BPE}}{\emph{GloVe-BPE}} achieves significantly higher scores than \emph{fastText} in almost all tests, \rev{}{which could be due to the segmenting mechanism enabled by the BPE that can capture the morphological boundaries better than n-gram based \emph{fasttext} given the highly fusional nature of Russian morphological marking}. 
        
        \rev{}{Our intrinsic experiment results show that the probing task performance and the improvement compared to the majority baseline differs depending on the language and the task, signaling that not all languages and morphological categories are equally easy to model. 
        There exist several ways to quantitatively capture morphological complexity, e.g. a recent work by \citet{cotterell2019complexity} plots the \emph{morphological counting complexity (MCC)} of the languages (defined as the number of cells in a language's inflectional morphological paradigm) against a novel entropy-based \emph{irregularity} measure to empirically demonstrate the hypothesized bound on the two complexity types: while a language can have a large paradigm or be highly irregular, it's never both. While paradigm-based counting complexity cannot be applied to the probing tests directly due to their categorical nature, one can use the number of unique values in a respective category as a rough approximation of the complexity of this category. For instance, a weak correspondence can be seen between the number of values and the baseline performances for \emph{Case} test- the less cases a language has, the higher the baseline. German with 4 cases have the majority baseline of 34.2, while Finnish with 15 cases have 30.0 as given in Table~\ref{tab:all-probing-results}. However, this pattern vanishes as we move to the embedding-based models: the end performance does not seem to depend on the number of case values, e.g., \emph{D-ELMo} performs equally well for Russian (6 cases) and Finnish (15 cases). This can indeed be related to the trade-off between the number of inflection paradigms and the irregularity, discussed by \namecite{cotterell2019complexity}. The regularity of the language (e.g., Finnish) may help the embedding models to learn the patterns and lead to even higher final scores than irregular languages (e.g., German, Russian) despite much lower baseline scores due to larger number of paradigms.}

      

        We observe that the static embeddings, \emph{word2vec} and \emph{MUSE}, which \rev{can not handle}{do not have a dedicated mechanism to flexibly represent} OOV words, performed similarly and had lower scores than other \rev{embeddings}{embedding models} for most of the tests, except from \textit{Pseudo}. Especially \emph{MUSE} has an outstanding performance on \textit{Pseudo} tests, compared to its performance on other tests. \rev{This is due to having an internal vocabulary and treating all words out of their dictionary the same, \textit{e.g., assigning the same random vector}, that leads to easier classification of non vocabulary words.}{This is not unexpected: in case of \textit{Pseudo}, the OOV handling mechanism of static embeddings, which maps unseen words to the same entry, puts static models at advantage since they encode all unknown words with a single random vector, making the detection of explicit out-of-vocabulary items easier.}
        \begin{table}[!ht]
            \centering
            \caption{Downstream tasks results for all languages. \textbf{Bold} represents the best score, while \textit{italics} is the second best.}
            \label{tab:all-ds-results}
            \scalebox{0.73}{
            \begin{tabular}{lccccc}
            \hline
            \multicolumn{6}{c}{\textit{Finnish}} \\
            \hline
            \toprule
            Task & MUSE & word2vec & \rev{BPE}{GloVe-BPE} & fastText & D-ELMo \\
            \midrule
            SRL &  62.30 & 57.68 & 60.41 & \textit{64.19} & \textbf{72.26} \\
            DEP & 79.62 & 79.84 & 80.6 & \textit{82.45} & \textbf{87.78} \\
            POS & 89.56 & 89.86 & 89.88 & \textit{92.55} & \textbf{96.56} \\
            NER & 72.96 & 71.17 & 75.69 & \textbf{80.54} & \textit{78.45} \\
            \bottomrule
            \hline
            \multicolumn{6}{c}{\textit{German}} \\
            \hline
            \toprule
            Task & MUSE & word2vec & \rev{BPE}{GloVe-BPE} & fastText & D-ELMo \\
            \midrule
            SRL &  55.25 & 60.60 & 57.11 & \textit{61.75} & \textbf{61.85} \\
            DEP & 82.43 & \textit{82.78} & 82.32 & 83.20 & \textbf{83.46} \\
            POS & 91.82 & 92.14 & 90.59 & \textit{92.66} & \textbf{93.57} \\
            NER & 74.32 & \textit{76.13} & 71.43 & \textbf{78.35} & 71.81 \\
            XNLI & 44.03 & 40.08 & 43.55 & \textbf{44.69} & \textit{44.05} \\
            \bottomrule
            \hline
            \multicolumn{6}{c}{\textit{Spanish}} \\
            \hline
            \toprule
            Task & MUSE & word2vec & \rev{BPE}{GloVe-BPE} & fastText & D-ELMo \\
            \midrule
            SRL &  64.49 & 62.78 & 62.34 & \textit{66.39} & \textbf{70.03} \\
            DEP & 90.17 & 90.26 & 89.99 & \textit{90.55} & \textbf{91.09} \\
            POS & 96.07 & \textit{96.58} & 95.66 & 96.49 & \textbf{97.43} \\
            NER & 77.48 & \textbf{79.31} & 77.36 & \textit{78.96} & 77.75 \\
            XNLI & \textit{46.75} & 41.28 & 45.17 & \textbf{46.80} & 45.07 \\
            \bottomrule
            \hline
            \multicolumn{6}{c}{\textit{Russian}} \\
            \hline
            \toprule
            Task & MUSE & word2vec & \rev{BPE}{GloVe-BPE} & fastText & D-ELMo \\
            \midrule
            DEP & 90.13 & \textit{90.54} & 90.16 & 87.41 & \textbf{92.26} \\
            POS & 95.62 & \textit{96.11} & 95.91 & 92.61 & \textbf{97.84} \\
            NER & 78.38 & \textbf{79.92} & 75.84 & 64.20 & \textit{79.71} \\
            XNLI & 43.43 & 39.80 & \textit{43.53} & 41.64 & \textbf{45.05} \\
            \bottomrule
            \hline
            \multicolumn{6}{c}{\textit{Turkish}} \\
            \hline
            \toprule
            Task & MUSE & word2vec & \rev{BPE}{GloVe-BPE} & fastText & D-ELMo \\
            \midrule
            SRL &  53.29 & 46.35 & \textit{53.51} & 53.14 & \textbf{63.38} \\
            DEP & \textit{57.82} & 56.67 & 55.92 & 57.70 & \textbf{62.97} \\
            POS & 86.52 & 87.35 & \textit{87.57} & 86.80 & \textbf{94.48} \\
            NER & 48.87 & \textit{52.21} & 51.75 & \textbf{52.52} & 49.22 \\
            XNLI & 42.79 & 42.93 & \textbf{45.17} & \textit{44.25} & 43.81 \\
            \bottomrule
            \end{tabular}
            }
        \end{table}
        
        We present the results of the extrinsic experiments in Table \ref{tab:all-ds-results}. The general performance ordering of the embeddings: \emph{D-ELMo}, \emph{fastText/\rev{BPE}{GloVe-BPE}}, \emph{word2vec/MUSE} holds for syntactic (POS, DEP) and shallow semantic tasks (SRL) for all languages, similar to the ranking in intrinsic experiments. However, for NER and XNLI tasks, we do not observe the same trend. \rev{}{There might be several reasons for this discrepancy. First of all, the extrinsic tasks at hand are conceptually different: while grammar-based POS, DEP and SRL directly build upon subword information (e.g. via agreement), for NER lexical content and surface cues play a bigger role, while XNLI as a semantic task benefits from lexical information and proposition-level cues like negation, tense and modality, rather than general subword-level phenomena. Hence, the lexical differences between the training corpora of multilingual embeddings and downstream tasks may have been more emphasized for these tasks, especially NER.~\footnote{Unlike the others, \emph{D-ELMO} has been induced from a \textit{subset} of Wikipedia: we hypothesize that it is enough to learn good representations for grammatical phenomena in common words, but not enough to populate the entity vocabulary. Given that most NER datasets are Wikipedia-based, this could lead to lower entity vocabulary intersection compared to the other embedding spaces, and thereby to lower scores. Testing this hypothesis, however, is not trivial without access to the exact source corpora and a reliable method to identify entities in them.}. Another potential reason for the difference in ranking is the domain of the data underlying the respective datasets: for the majority of the languages, POS, DEP and SRL data originates from the same treebanks and has gold (expert) annotations. On the other hand, NER and XNLI datasets are generally compiled from a different, and often diverse set of resources. Third reason may be the different out-of-vocabulary (OOV) ratios among different datasets. In order to investigate this, we have calculated the OOV ratio of development and test sets of each extrinsic task with respect to the training set, shown in Table~\ref{tab:extrinsic-data-stat}. We observe that XNLI task has the lowest OOV ratio among all other extrinsic tasks for all languages. Similarly, when OOV ratios of extrinsic tasks with respect to our static embeddings (MUSE and word2vec) are examined (shown in Appendix~\ref{app:oov-rate}, Table~\ref{tab:muse-word2vec-extrinsic-oov}), we notice that both embeddings have the lowest OOV ratio for XNLI task. These statistics could indeed explain the smaller gaps between static and subword-level models for the XNLI task. Finally, NER annotations for most of the experimented languages are of silver quality, i.e., there exists many incorrect and missing labels; and the multilingual sentences provided in XNLI are automatically translated by an existing tool.}

        \rev{}{We observe that static word embedding spaces (\emph{word2vec} and \emph{MUSE}) rank generally higher on downstream tasks compared to the probing tasks for fusional languages (German, Spanish, Russian). We attribute this to the vocabulary difference between the extrinsic and intrinsic datasets. As mentioned in Sec.~\ref{ssect:testcreation}, our type-level probing data contains many word forms that rarely occur in Wikipedia main text, which is the primary text source for the vector space models we compare. The extrinsic datasets, on the contrary, are derived from Wikipedia and newswire, resulting in a higher lexical overlap, lower unseen word rate, and therefore better performance. We have calculated the OOV rates of intrinsic and extrinsic tasks, relative to both word2vec and MUSE embeddings and found that OOV rates for our probing tasks are indeed much higher than our extrinsic tasks supporting our hypothesis. The OOV rates are given in Appendix~\ref{app:oov-rate}.}
        

    
        \subsubsection{\secrev{}{Results on Token-Level Probing Tasks}}
        \label{ssec:analysis_token_level}

        \secrev{}{In order to investigate the token-level probing tasks even more deeply, we apply the same experimental setup described in Sec.~\ref{ssec:intrinsic} to the token-level probing test suite and present the results in Table~\ref{tab:token-level-probing-results}. Since the majority of the embedding spaces used in this study~(see Sec.~\ref{ssec:embeds}) are not contextualized, \textit{i.e., have the same representation for the surface form independent from its surrounding words}, we only use the token itself without its context. That means, when tokens are isolated in such a way, there may be duplicates among training, development and test sets. Therefore the results for MUSE, word2vec, GloVe-BPE and fastText are only provided for comparison among each other; and to gain insights on some of the aspects discussed in Sec.~\ref{ssec:disc_prob}. Finally to provide a more realistic use-case for the token-level probing task, we experiment with the ELMo embeddings without decontextualizing them, referred to as contextualized ELMo (C-ELMo).}  

        \secrev{}{To categorize our findings for token/type-level probing tasks, we use some of the aspects from our previous discussion in Sec.~\ref{ssec:disc_prob}. 
        \paragraph{Dataset size} We observe that having a smaller dataset size for Turkish token-level probing tasks eliminated the possibility to probe for the ``Possession" feature. We suspect that it may occur for other relatively low-resourced languages, leaving us with only the most common, generic tasks. 
        \paragraph{Tag Frequency} As discussed previously, constructing tasks on an annotated corpora may introduce biases towards frequently encountered feature values in the dataset. When the gap between the majority baseline scores is examined, it can be seen that for certain features, e.g., Mood and Number, the gap is in the range of 40-60\%. For instance ``Polarity" feature for Turkish has 89\% majority voting score, meaning that 89\% of the instances had the ``Positive" label. \paragraph{Token Frequency} Earlier we have hypothesized a token frequency bias when using a full-text based probing test. When we compare the performance gaps of embedding models between type and token-level tasks, we observe a substantial performance boost for the static embeddings: MUSE and word2vec, for all languages and tasks apart from few exceptions; while we observe smaller gaps or performance drops for subword-based dynamic models: bpe, fasttext and D-ELMo. The performance boost of static embeddings is again related to lower OOV ratio in the token-level datasets. 
        \paragraph{Lexical Variety} This effect is visible from the results of the ``POS" feature, compared to type-level ``POS" test. First, the token-level POS baseline scores are noticeably lower; and second, all embedding spaces including D-ELMO achieve much lower scores. \paragraph{Ambiguity} Token-level probing unlocks a few more probing tasks such as ``Gender" for German, for which we did not have access before due to eliminating ambiguous forms. More importantly, we observe that removing ambiguous forms may have introduced a sort of bias towards some of the features, i.e., simplified the task by eliminating certain feature values that always produce ambiguous surface form. This effect can be easily observed the performance gaps between D-ELMo and C-ELMo. In other words, when a certain feature gets a performance boost by C-ELMo, this may suggest that the feature is highly ambiguous and a model that can use contextual information to resolve ambiguity outperforms the one that can't by a large margin. The following feature-language pairs demonstrate the described phenomena: German-Case, German-Number, German-Gender and Russian-Case; which had the highest ambiguity ratios as discussed in Sec.~\ref{ssec:disc_prob}. Apart from these cases, we see a similar pattern for the ``POS" feature; however this feature is also affected by the limited lexical variety of type-level tasks. Therefore, there are two phenomena responsible for the performance boost for ``POS".}   

        \secrev{}{Finally, when embeddings are compared amongst each other (except from C-ELMo due to having many duplicate training and test instances in token-level tests when tokens are isolated from context), we see a similar ranking for each language-feature pair, suggesting that type- and token-level probing tasks have many commonalities despite their differences discussed above.} 

        \begin{table}[!ht]
          \centering
          \caption{\rev{}{Token-Level probing task results for all languages. \textbf{Bold} represents the best score, while \textit{italics} is the second best.}}
          \label{tab:token-level-probing-results}
          \scalebox{0.7}{
          \begin{tabular}{lccccccc}
          \hline
          \multicolumn{8}{c}{\textit{Finnish}} \\
          \hline
          \toprule
          Task  &  baseline  &  MUSE  &  word2vec  &  GloVe-BPE  &  fastText  &  D-ELMo  &  C-ELMo \\
          \midrule
          Case  &  35.9 & 79.7 & 69.1 & 85.0 & 97.3 & \textit{97.9} & \textbf{98} \\
          Mood  &  89.7 & 94.9 & 94.3 & 96.7 & \textit{97.4} & 97.3 & \textbf{98.3} \\
          Number  &  82.2 & 92.8 & 90.4 & 94.4 & 97.8 & \textit{98.3} & \textbf{98.7}\\
          POS  &  29.5 & 69.0 & 72.0 & 68.9 & 71.3 & \textit{74.8} & \textbf{87.5} \\
          Person  &  64.5 & 92.8 & 89.7 & 95.1 & \textbf{97.3} & \textit{96.8} & 96.4 \\
          Tense  &  62.9 & 95.2 & 92.8 & 97.5 & \textbf{98.4} & \textbf{98.4} & \textit{98.1} \\
          Voice  &  86.2 & 96.2 & 92.7 & 96.5 & \textit{97.7} & \textbf{98.0} & 97.5 \\
          \bottomrule
          \hline
          \multicolumn{8}{c}{\textit{German}} \\
          \hline
          \toprule
          Task  &  baseline  &  MUSE  &  word2vec  &  GloVe-BPE  &  fastText  &  D-ELMo  &  C-ELMo \\
          \midrule
          Case  &  32.7 & 56.9 & 52.9 & 53.0 & \textit{56.8} & 53.4 & \textbf{80.8} \\
          Gender  &  38.6 & 72.6 & 69.3 & 66.5 & \textit{73.5} & 72.8 & \textbf{78.7} \\
          Mood  &  96.4 & 98.3 & 98.4 & 98.4 & \textbf{98.9} & \textit{98.7} & \textbf{98.9} \\
          Number  &  79.5 & 88.5 & 88.1 & 86.1 & \textit{89.5} & 89.2 & \textbf{94.7} \\
          POS  &  20.6 & 76.5 & 82.6 & 74.5 & 76.4 & \textit{77.1} & \textbf{91.2} \\
          Person  &  72.8 & 94.6 & 94.1 & 93.8 & \textit{95.1} & \textit{95.1} & \textbf{97.9} \\
          Tense  &  51.6 & 98.7 & 97.8 & 98.7 & 98.6 & \textbf{99.2} & \textit{98.8} \\
          \bottomrule
          \hline
          \multicolumn{8}{c}{\textit{Spanish}} \\
          \hline
          \toprule
          Task  &  baseline  &  MUSE  &  word2vec  &  GloVe-BPE  &  fastText  &  D-ELMo  &  C-ELMo \\
          \midrule
          Gender  &  58.0 & 98.6 & 95.3 & 97.9 & 99.2 & \textit{99.3} & \textbf{99.4} \\
          Mood  &  92.3 & 97.7 & 97.2 & 95.9 & \textbf{99.1} & \textit{98.2} & 97.8 \\
          Number  &  75.2 & 98.7 & 96.4 & 98.6 & \textbf{99.6} & \textbf{99.6} & \textbf{99.6} \\
          POS  &  19.3 & 78.1 & 83.2 & 78.0 & \textit{79.6} & 78.9 & \textbf{92} \\
          Person  &  68.4 & 97.4 & 95.3 & 98.1 & \textbf{99.2} & \textit{99.0} & \textit{99.0} \\
          Tense  &  55.4 & 98.2 & 97.1 & 98.7 & \textbf{99.3} & \textit{99.2} & 99.1 \\
          \bottomrule
          \hline
          \multicolumn{8}{c}{\textit{Russian}} \\
          \hline
          \toprule
          Task  &  baseline  &  MUSE  &  word2vec  &  GloVe-BPE  &  fastText  &  D-ELMo  &  C-ELMo \\
          \midrule
          Case  &  31.7 & 75.3 & 47.6 & 68.2 & 78.6 & \textit{78.7} & \textbf{89.7} \\
          Gender  &  45.4 & 86.0 & 64.7 & 87.0 & \textbf{89.9} & \textit{89.8} & 89.7 \\
          Number  &  75.3 & 93.9 & 78.2 & 92.8 & 96.8 & \textit{97.0} & \textbf{97.4} \\
          POS  &  30.3 & 76.4 & 64.5 & 73.3 & 76.5 & \textit{76.9} & \textbf{95.4} \\
          Person  &  51.1 & 91.8 & 84.3 & 98.1 & \textit{98.6} & \textit{98.6} & \textit{98.8} \\
          Tense  &  53.0 & 90.6 & 78.4 & \textit{97.6} & \textbf{98.7} & 97.2 & 96.9 \\
          Voice  &  69.5 & 88.9 & 79.0 & \textbf{95.6} & 93.7 & \textit{95.4} & 94.5 \\
          \bottomrule
          \hline
          \multicolumn{8}{c}{\textit{Turkish}} \\
          \hline
          \toprule
          Task  &  baseline  &  MUSE  &  word2vec  &  GloVe-BPE  &  fastText  &  D-ELMo  &  C-ELMo \\
          \midrule
          Case  &  47.7 & 84.3 & 67.7 & 89.0 & 91.4 & \textit{95.4} & \textbf{96.6} \\
          POS  &  35.4 & 71.7 & 68.0 & 75.4 & \textit{78.3} & 77.4 & \textbf{83.2} \\
          Person  &  78.3 & 91.0 & 84.3 & 95.0 & \textbf{97.2} & \textit{96.8} & 96.5 \\
          Polarity  &  89.0 & 95.4 & 92.3 & 98.3 & \textbf{99.2} & \textit{99.0} & 98.8\\
          Tense  &  55.2 & 87.7 & 76.3 & 94.6 & \textbf{96.6} & 95.9 & \textit{96.3} \\
          \bottomrule
          \end{tabular}
          }
        \end{table}  

\section{Analysis}
In this section we investigate the relation between downstream and the probing tasks more closely, and report the results with respect to language families and downstream tasks. We present the results for the diagnostic case study described in Sec.~\ref{ssec:diagtool} and show the close connection to highly correlated probing tests. Finally we give a brief summary of our findings related to proposed probing tasks. 

\subsection{Correlation}
\label{ssect:correl}

\begin{figure*}
  \centering
    \begin{subfigure}[b]{0.45\textwidth}
      \includegraphics[width=\textwidth]{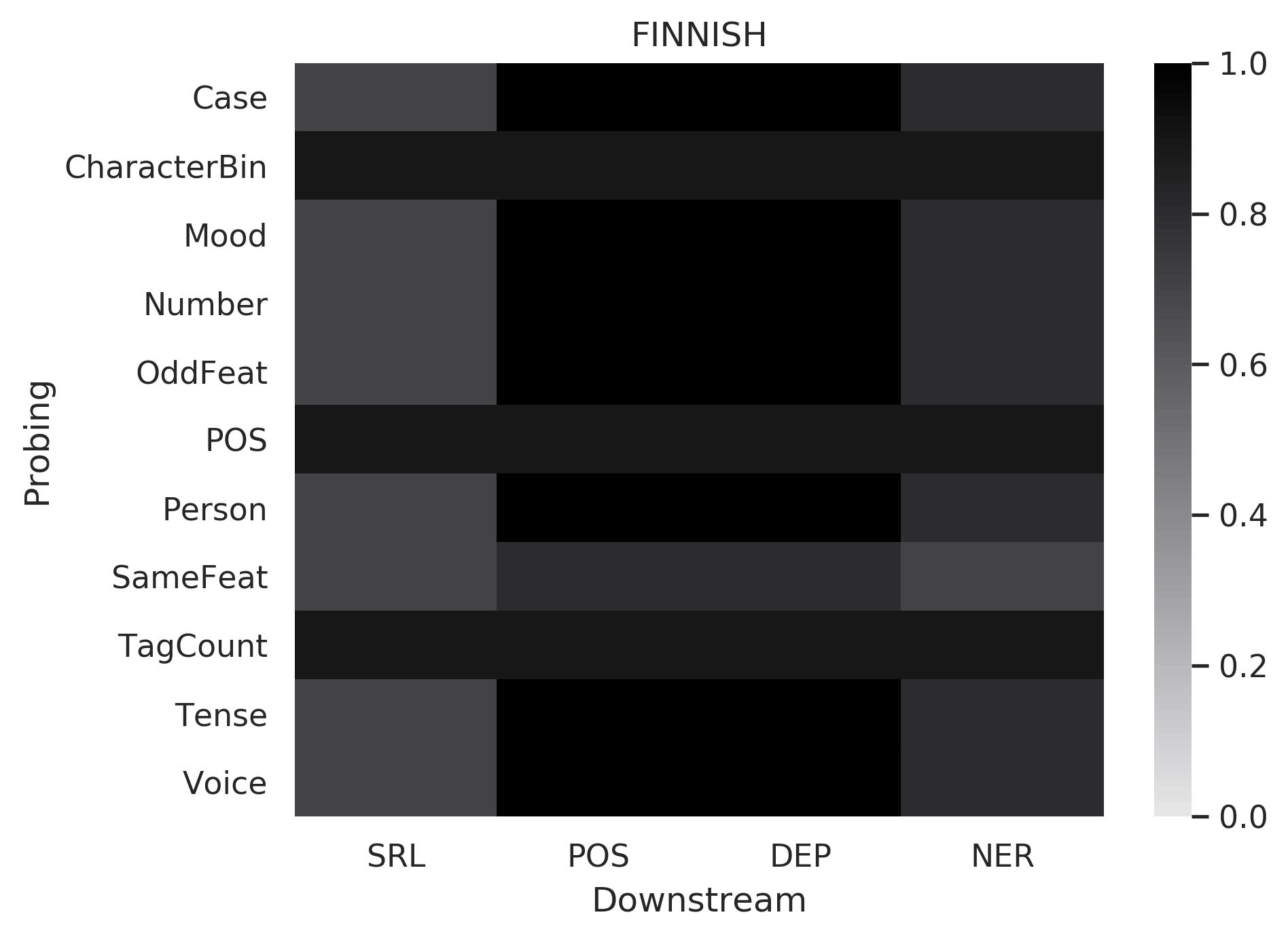}  
  \end{subfigure}
  ~ 
  \begin{subfigure}[b]{0.45\textwidth}
      \includegraphics[width=\textwidth]{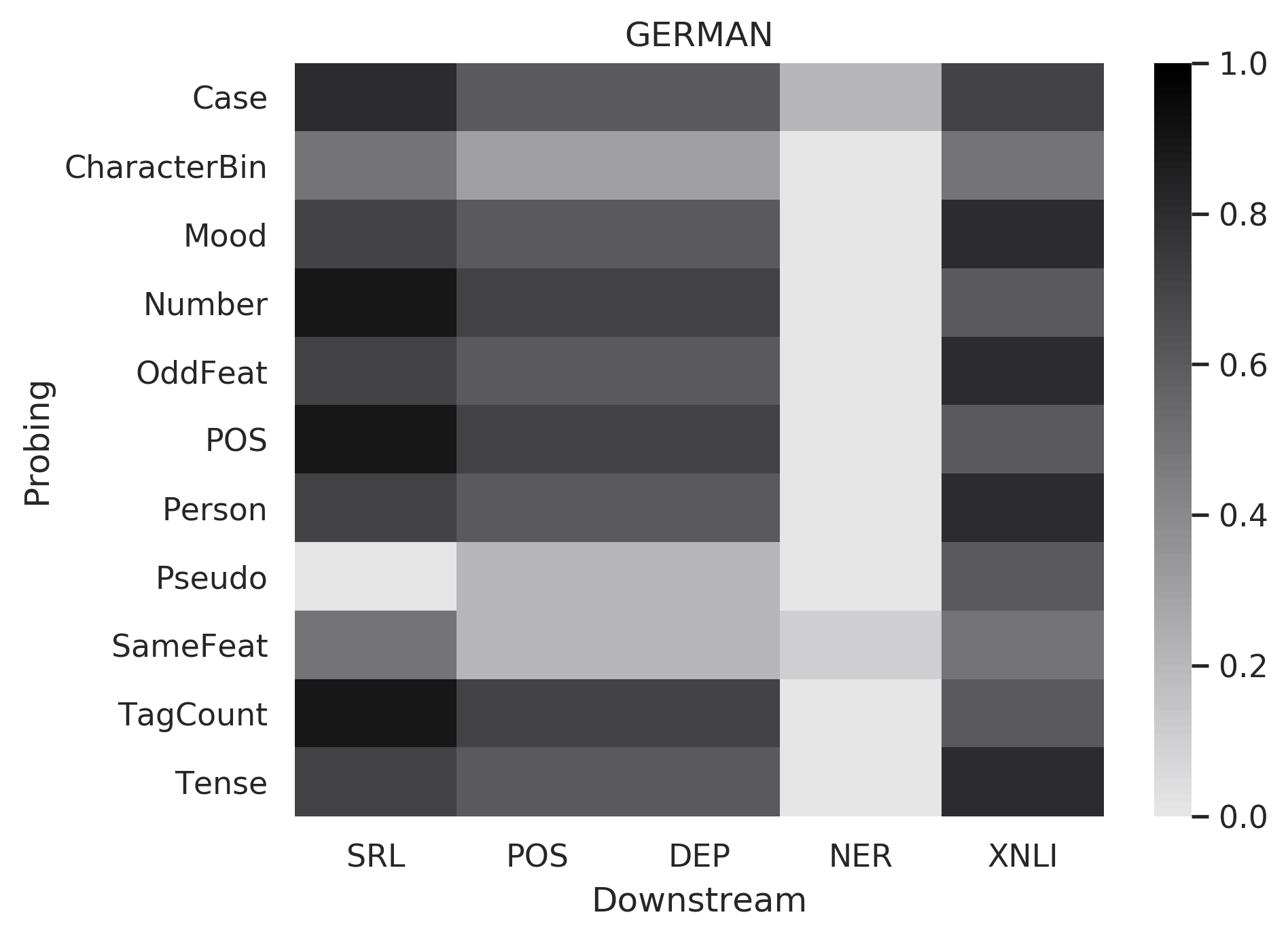} 
  \end{subfigure}
  ~ 
  \begin{subfigure}[b]{0.45\textwidth}
      \includegraphics[width=\textwidth]{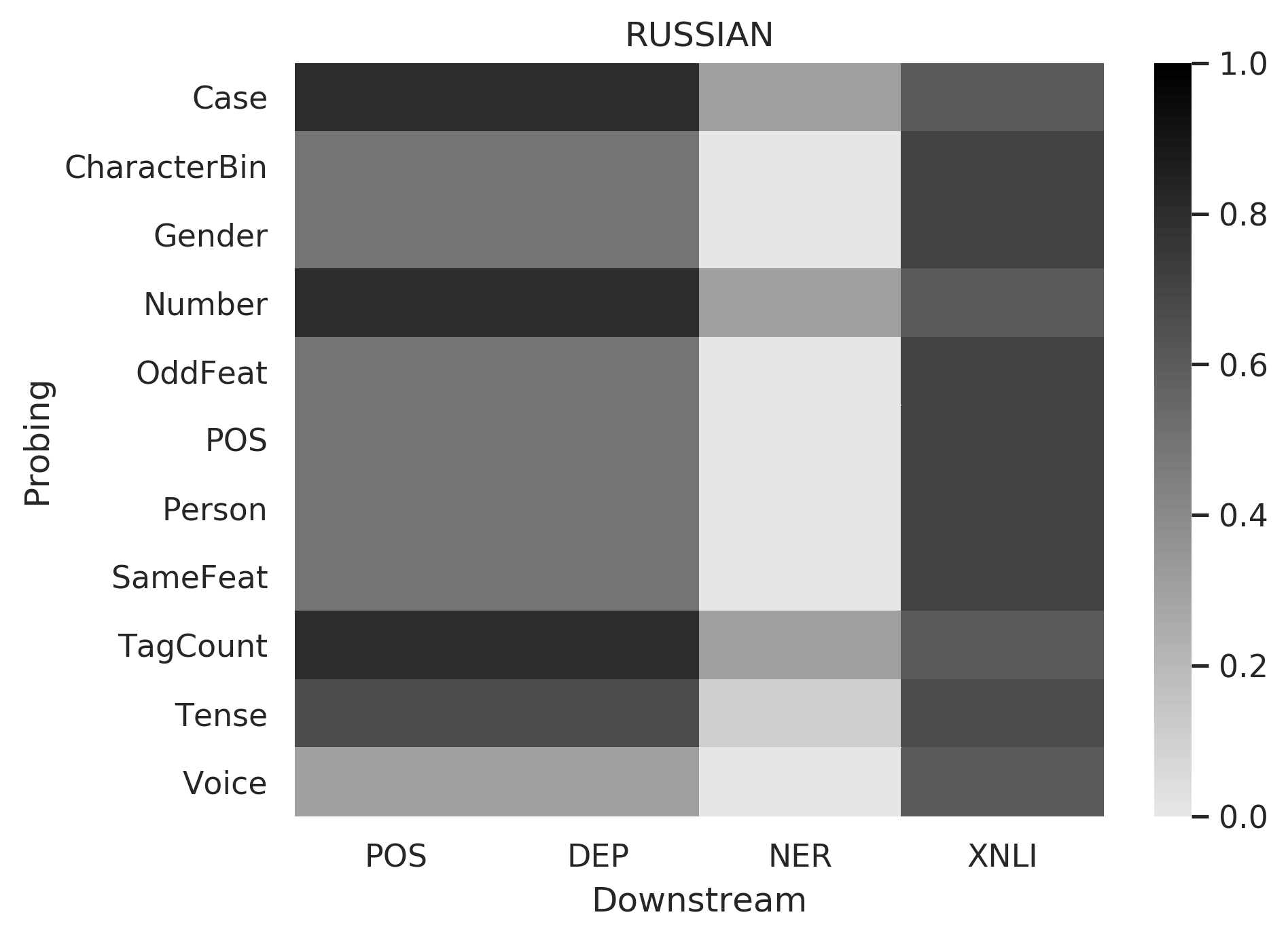}
  \end{subfigure}
  ~ 
  \begin{subfigure}[b]{0.45\textwidth}
      \includegraphics[width=\textwidth]{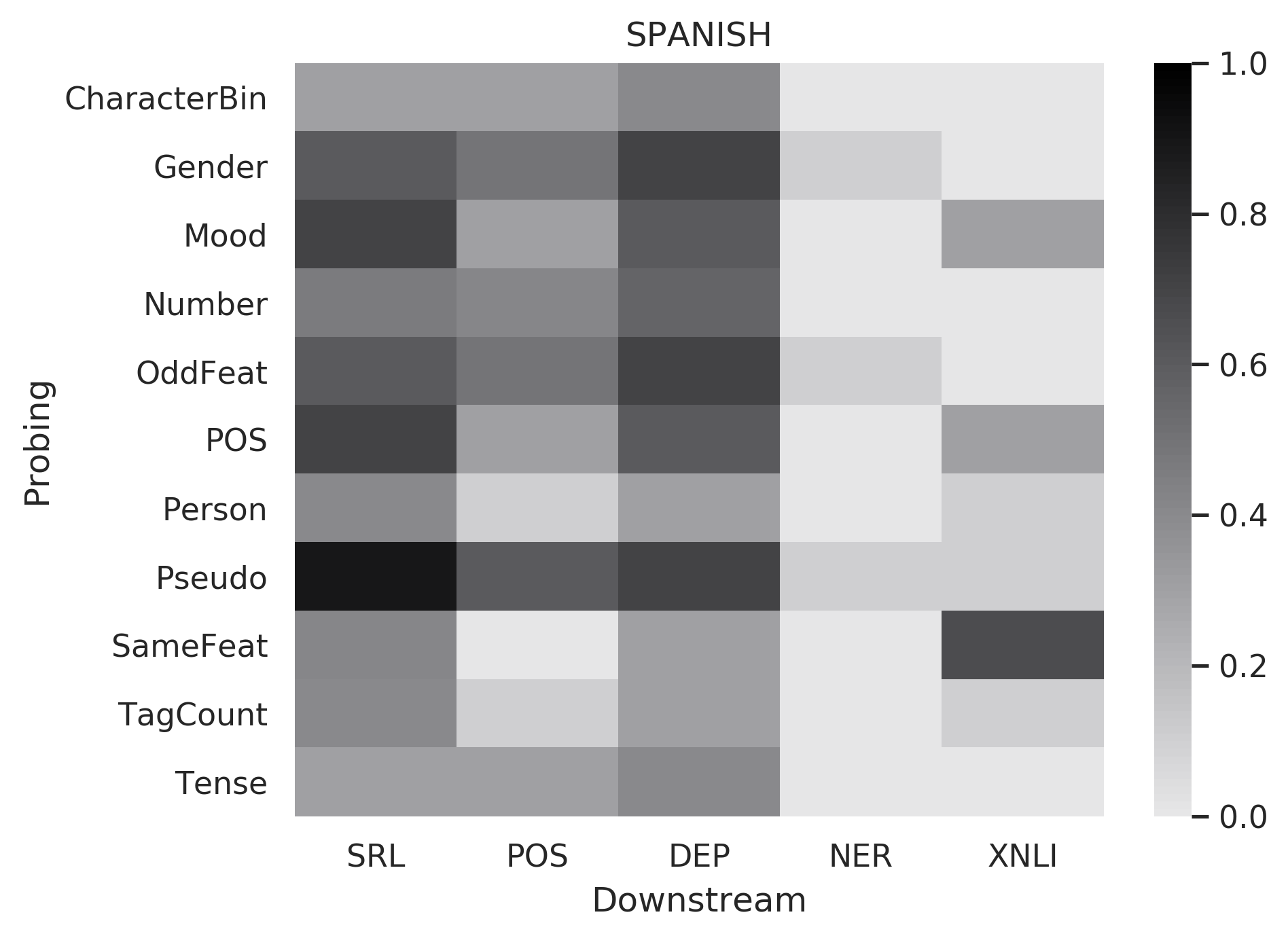}
  \end{subfigure}
  ~ 
  \begin{subfigure}[b]{0.45\textwidth}
      \includegraphics[width=\textwidth]{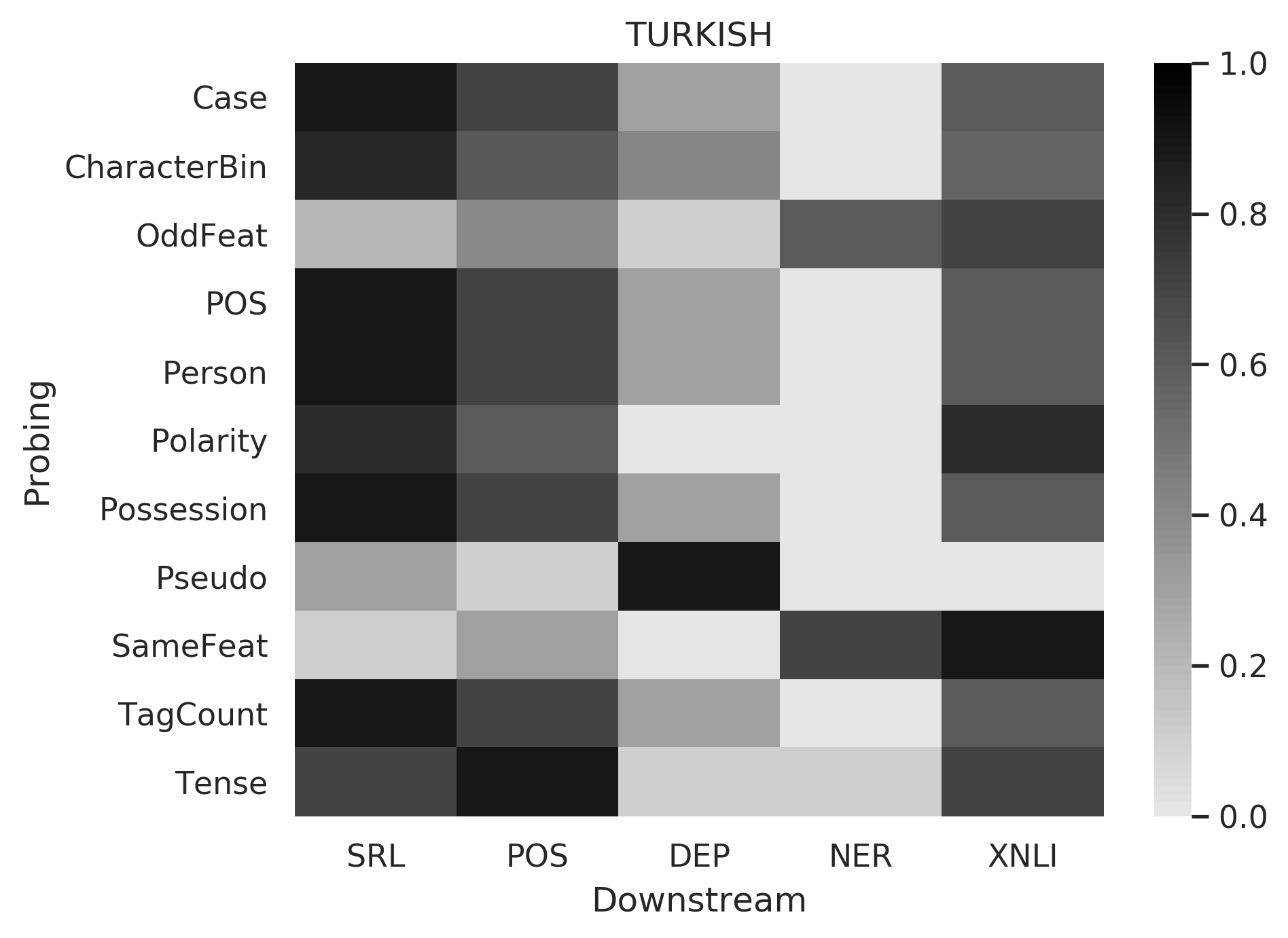}
  \end{subfigure}
  \caption{Spearman correlation between probing and downstream tasks for each language.}
  \label{fig:spearman}
  \end{figure*}

  In order to calculate the relation between the downstream tasks and the probing tests, we calculate the Spearman correlation coefficient as shown in Fig.~\ref{fig:spearman}. In addition, we calculate the two-sided \emph{p-values} to test the null hypothesis, i.e., whether two sets of results are uncorrelated, and interpret the results with respect to the languages and the tasks.~\footnote{Significant correlations are given in Appendices.} 

  \subsubsection{Language-related findings} 
  
  \paragraph{Finnish} We observed the highest correlations with \emph{p-value} of $0.1$ in Finnish language~\footnote{Since the number of samples, i.e., number of embeddings, for the correlation analysis are 5, we use a high p-value of 0.1.}. According to the calculated \emph{p-values}, all proposed tests, except from \textit{SameFeat}, had a statistically significant correlation with POS, DEP, SRL and NER for Finnish. \rev{}{As already shown in Table~\ref{tab:uni_versus_ud}, in type-level statistics columns, Finnish data had the lowest ambiguity ratio, highest number of surface forms and paradigms, and the highest lexical diversity, that leads to strong correlations to downstream tasks. Furthermore, being an agglutinative language with high morphological counting complexity (MCC) but lower irregularity, it allows encoding considerable amount of syntactic-semantic information on type-level, which is another explanation for strong correlations of the proposed tasks.}
  
  \paragraph{Turkish} For Turkish, we found strong correlations for all single feature tests for syntactic tasks (except from DEP), and registered relatively high correlation between \textit{Polarity}, \textit{SameFeat}, \textit{OddFeat}, \textit{Tense} and XNLI task. \rev{}{Although Turkish is typologically similar to Finnish and has a similar degree of MCC and irregularity, we observe weaker correlations for POS and inconclusive correlations for DEP. According to Table~\ref{tab:uni_versus_ud}, the data that Turkish probing tasks are originated from, are only around 10\% of Finnish data with a slightly higher ambiguity ratio. Although the lexical variety in terms of word classes is similar, smaller data size and more importantly smaller number of paradigms, i.e., less variety encoded on type-level, may have influenced the correlation scores negatively. Finally, the respective treebanks of both languages have been sourced from different domains. While Finnish treebank is based on a grammar book, Turkish data is a combination of domains ranging from news data to children stories. It wouldn't be unexpected to have higher correlations between tasks from similar domains as in Finnish.}
  
  \paragraph{German} For German, we have observed high correlation with \emph{p-value} of $0.1$ for \textit{Number}, \textit{POS} and \textit{TagCount} tests, whereas \textit{Case}, \textit{Mood}, \textit{OddFeat}, \textit{Person} and \textit{Tense} have statistically significant correlation with \emph{p-value} of $0.2$ for SRL. For German, the correlation pattern of \textit{Case}, \textit{Number}, \textit{POS} and \textit{TagCount} repeated for syntactic and shallow semantic tasks: POS, DEP and SRL, whereas XNLI correlated well with \textit{Case}, \textit{Mood}, \textit{OddFeat}, \textit{Person} and \textit{Tense}. \rev{}{We observe that the correlating tasks are similar to those of agglutinative languages in general (except from \emph{CharacterBin} -- explained below), however weaker. The weaker correlations may be the result of highly ambiguous nature of German data (especially \emph{Case} and \emph{POS}), and less lexical diversity, \emph{that are both common among fusional languages}, as shown in Table~\ref{tab:uni_versus_ud}. In addition, the paradigm sizes for German nouns and verbs are only 29 and 8, (for reference Turkish has 120 and 100 respectively)~\cite{CotterellMER18}.} 

  \paragraph{Russian} For Russian, we find that \textit{Case}, \textit{Number} and \textit{TagCount} to have high correlations with \emph{p-value} of $0.2$ for syntactic tasks, \rev{}{similar to German}, whereas XNLI correlated better with the other features such as \textit{SameFeat}, \textit{OddFeat}, \textit{Person} and \textit{Tense}. \rev{}{Russian is amongst fusional languages like German; with similar MCC and irregularity values. One exception is the high irregularity of Russian verbs (with a score of 1.67) compared to German verbs that have a score of 0.77~\cite{CotterellMER18}. This could explain the weaker correlations for verb-related probing tasks such as \textit{Person} and \textit{Tense} as can be seen more clearly in Fig.~\ref{fig:spearman}.}

  \paragraph{Spanish} For Spanish, there was no clear correlation pattern, except from the \textit{Pseudo} test that had a strong correlation to SRL and DEP extrinsic tasks with $p=0.2$. \rev{}{Lack of correlations can be attributed to the lack of lexical variety in Spanish probing tasks discussed in Sec.~\ref{ssec:disc_prob}. In addition, Spanish is one of the languages with the highest gap of OOV ratio between its extrinsic and intrinsic datasets. For instance, MUSE has the OOV ratio of 2.71\% in training split of Spanish dependency treebank, while having minimum of 48\% for the probing tasks as given in Appendix~\ref{app:oov-rate}. This could lead to unnatural performance gaps between static and other embeddings, that can effect the correlation. }

  \paragraph{Summary} We observed that a large number of probing tasks had high correlation to syntactic tasks especially for \emph{agglutinative} languages: Turkish and Finnish. \rev{}{This result can be connected to several of our previous observations discussed in Sec.~\ref{ssec:disc_prob}, namely regarding the \emph{lexical variety} and the \emph{the ambiguity ratio}. In Sec.~\ref{ssec:disc_prob}, we have demonstrated that the probing tests for these languages had the richest lexical variety with the coverage of nouns, verbs and adjectives; and had the lowest ambiguity ratio mostly due to the property of one morpheme encoding only one morphological feature. These observations suggest that the instances in the proposed tests are satisfactory representatives of the language. Apart from these observations, another reason is the amount of information encoded by morphological features on the type-level, hinted by the regularity of the languages~\cite{cotterell2019complexity}. As shown in~\citet{CotterellMER18}, Finnish and Turkish have much lower irregularity scores compared to the other languages we have experimented with. It can be interpreted as, the more regularity a language has, the more information can be incorporated into a ``single unit/word'' without introducing more ambiguity, which makes it easier to capture more syntactic information in a single form.} 

  On the other hand, we found a smaller set of probing tasks with high correlation for \emph{fusional} languages, especially for syntactic tasks. \rev{}{This can be connected to the amount of syntactic information that can be encoded in a word, which can be linked to the number of paradigms. For instance, while Turkish has 120 noun paradigms, German and Russian only have 29 and 25 respectively~\cite{CotterellMER18}, suggesting that the amount of information encoded in nouns are indeed limited. In addition, lack of \emph{lexical variety} (having only verbs) has been observed to have the biggest impact on Spanish which caused the weakest correlations among all languages.} Furthermore, we observed that a set of common probing tests have higher correlation to certain downstream tasks among most languages, such as \textit{Case}, \textit{POS}, \textit{Number} and \textit{TagCount} to syntactic tasks; and \textit{OddFeat}, \textit{SameFeat}, \textit{Tense} to XNLI\rev{.}{; while we haven't detected any strong correlation to NER for almost all languages.} \rev{}{Case is among the features that signals syntactic and semantic connection between nominals and verbs as discussed in Sec.~\ref{ssect:tesdefs}; hence it has been expected to correlate well with syntactic and shallow semantic tasks. \textit{POS} is an obvious syntactic feature, while \textit{Number} is not. However, \textit{Number} is among the common grammatical agreement features and provides clues for linking syntactically related words that should agree on number, (e.g., linking subject and verb in dependency parsing). The result of the \textit{TagCount} test suggests that simplistic approximation of morphology may be a good predictor for syntactic task performance.} \rev{}{We also note that there are cases where a correlation was hypothesized but never observed or has been found weak, such as simple morphological feature probing tasks, e.g., \emph{Case}, \emph{Tense}, \emph{Number}, and Turkish dependency parsing. Similarly, in some cases, a correlation is found although not hypothesized such as Turkish NER correlating well with \emph{SameFeat} feature. These suggest that there is a certain amount of noise in the correlation measurements that may be a result of many different factors such as small number of data points (embedding models). While our results reveal certain patterns, obtaining the data points (language - extrinsic score - probing score) is computationally expensive , which limits the precision of the correlation tests.}
  
  \subsubsection{Downstream task-related findings}
  \label{ssec:downs_findings} 

  \paragraph{SRL} For all languages, SRL is found to correlate with the highest amount of probing tasks. This finding is intuitive since SRL performance is dependent on more complex linguistic phenomena compared to other tasks. Regardless of the languages families, we find that SRL has high correlations with \textit{Case} and \textit{POS}, generally followed by \textit{Person} and \textit{Tense} tests. This finding is on par with the traditional language independent features used for SRL back in the feature-engineering days~\cite{conll09}. In addition to those tests, for agglutinative languages, we find high correlation for \textit{CharacterBin} and \textit{TagCount} \rev{.}{(explained later in this section).} In addition, SRL has high correlation to \textit{Possession} and \textit{Polarity}, which only exists for Turkish. \rev{}{As discussed in Sec.~\ref{ssect:tesdefs}, \textit{Possession} provides a link between the possession of the object and person of the verb. This is highly relevant to SRL, which aims to detect the arguments of the predicates, hence it may help especially for the argument identification subtask. \textit{Polarity} can be directly linked to SRL due to having labeled negation arguments (\textit{ArgM-NEG}).} We see that \textit{Mood} is a common highly correlated test for fusional languages, whereas \textit{Number} only correlates with German SRL, and \textit{Pseudo} only correlates for Spanish SRL. \rev{}{\textit{Mood} can be considered highly relevant to SRL for several reasons. First, an imperative predicate generally implies that the subject (which usually undertakes the Agent or Patient role) is missing. Furthermore, other Mood values such as \emph{conditional} provide valuable links between the main predicate and the dependent clause, which is also labeled as an argument \textit{(ArgM-PRD)}.}

  \paragraph{POS and DEP} They can be considered easier tasks compared to SRL, where superficial linguistic cues would be enough to decide on local classes. However these cues are not expected to be distinct from SRL, rather a subset of it. Confirming this, for German we see that the set of highly correlated features are reduced to the subset: \textit{Case}, \textit{Number}, \textit{POS} and \textit{TagCount}. Another hint to support this hypothesis, is the decline in the correlation scores of \textit{CharacterBin} and \textit{TagCount} in POS and DEP for agglutinative languages. This finding suggests that instead of a feature that distantly approximates the morphological features of a given word such as \textit{TagCount}, a feature focused on a single linguistic phenomenon has higher correlation with more syntactic tasks. 

  \paragraph{NER} Except from Finnish NER, none of NER tasks had significantly high correlations to our probing tests. \rev{It should be noted that NER as a task is very different from POS, DEP and SRL. For instance, while morphological cues are of great importance to POS, DEP and SRL; for NER, having good representations for the \emph{entities} is more important. This finding suggests that either (1) the proposed probing tasks, including the Pseudo test, were not capable of capturing this phenomenon; or (2) the chosen embeddings were similarly syntactic and hence had a low variance.}{While POS, DEP and SRL represent different levels of grammatical analysis which are correlated with morphological phenomena, NER is a surface-level semantic task for which the lexical content of the target and surrounding tokens is by far more important than the morphological markers evaluated by our probing tests. The observed correlations to morphological probing tests are therefore weak and irregular among the languages.} 
  \paragraph{XNLI} For XNLI, we observe a noticeable pattern consistently among almost all languages, which is high correlation to \textit{Mood}, \textit{Polarity}, \textit{Tense} and \textit{Person} that \rev{had been hypothesized by~\namecite{N18-1101}}{is in aggreement with the original study by~\namecite{N18-1101}}; and high correlation with one of our paired tests (usually \textit{SameFeat}) that resembles the NLI task\rev{.}{, in a way that both tasks aim to capture the commonalities between a pair of tokens. However our probing tasks mostly capture morphological commonalities and differences, which might only constitute a subset of the phenomena relevant for NLI. As discussed in Section~\ref{ssec:results}, the largest overlap between the vocabulary of static embeddings is to the vocabulary of XNLI task, that leads to the low OOV ratios and smaller performance gaps between static and subword-level models for the XNLI task. This may have caused an unfair shift in rankings and, subsequently, the correlations, which deserves a separate dedicated study.}

  Furthermore, we notice that \textit{CharacterBin} and \textit{TagCount} are redundant tests for agglutinative languages. This is due to these languages having one to one morpheme/tag mapping, which suggests that the number of characters is also a good indicator for number of tags. Since other languages are fusional, i.e., exhibit a one to many relation between morphemes and tags, these tests do not relate to each other as can be seen from the correlation matrices of German and Russian, for which the \textit{TagCount} has high correlation scores unlike the \textit{CharacterBin}.

\subsection{Diagnostic Task}
\label{ssec:diags}

  In this section, we demonstrate the results and analysis of the diagnostic case study for Finnish and Turkish. As described in Sec.~\ref{ssec:diagtool}, we train an end-to-end SRL model which only uses character trigrams as input. We first probe the word encoding layer with the suggested type-level probing tests for three consecutive epochs, where we see a large improvement in SRL F1 scores. We probe Finnish for epochs: $2$, $8$ and $20$, which had F1 scores of $29.60$, $46.64$ and $57.93$. Similarly we probe the encoding layer of Turkish SRL for the epochs $4$, $6$ and $14$, with $15.55$, $43.06$ and $54.11$ F1 scores. Then we use the token-level tasks to diagnose the encoding layer together with the consequent LSTM layer.
  
  \subsubsection{Type-Level Diagnosis}
  \label{ssec:diags}
\begin{figure*}[!ht]
      \centering
        \begin{subfigure}[b]{0.75\textwidth}
          \includegraphics[width=\textwidth]{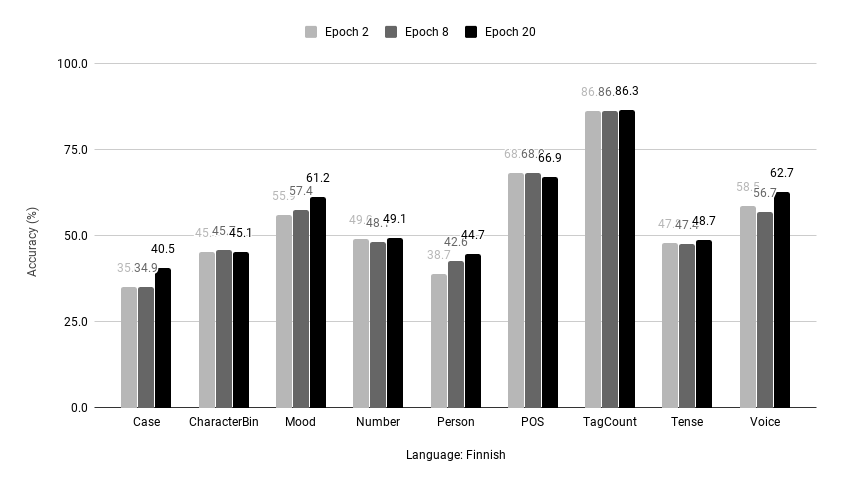}  
      \end{subfigure}
    
      \begin{subfigure}[b]{0.75\textwidth}
          \includegraphics[width=\textwidth]{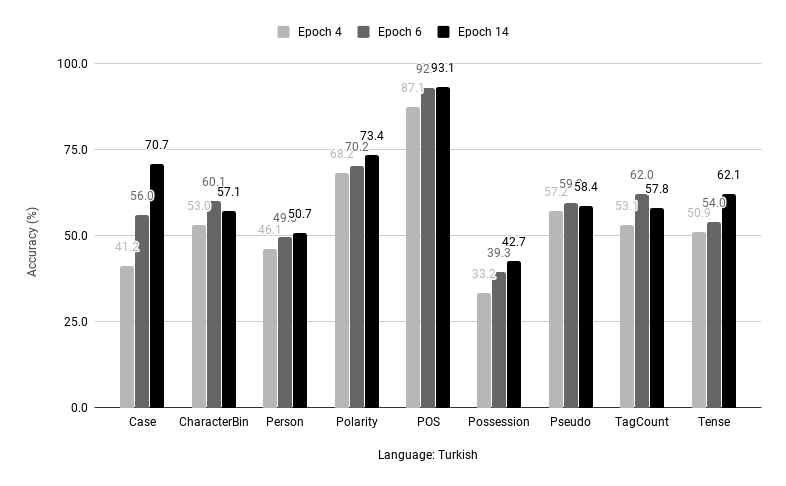} 
      \end{subfigure}
      \caption{Type-level probing tests to diagnose encoding layer of pretrained Turkish and Finnish SRL models}
      \label{fig:diagnostic_eval}
      \end{figure*}

      In case of Finnish, we have seen large improvements on \textit{Case}, \textit{Mood}, \textit{Person} and \textit{Voice}, while seeing a drop or a constant score for the features \textit{CharacterBin}, \textit{Number}, \textit{POS} and \textit{TagCount} as shown in Fig.~\ref{fig:diagnostic_eval}. These results suggest that the encoding layer captures more of \textit{Case}, \textit{Mood}, \textit{Person} and \textit{Voice} information throughout the training for an SRL objective -- these probing tests are also found to have significantly high correlation to Finnish SRL in our correlation study. 
      Interestingly, we observe constant or lower scores in correlated features such as \textit{CharacterBin} and \textit{Number}. We note that, even if these tests provide a predictive performance on the SRL task, not all neural models are capable of learning all correlated features discussed in previous section. This could be due to the lack of capacity of the neural model, or these features getting captured easily during the very early stages of the training. Since the aim of this section is to demonstrate a case study, a thorough comparison and in-depth investigation of the root causes is not in scope of this work. 
      
      Since the F1 improvements are more pronounced for Turkish, we see a more clear pattern in the probing task improvements. Similar to our results for Finnish, we have encountered considerably high improvements for \textit{Case}, \textit{Person}, \textit{Polarity}, \textit{POS}, \textit{Possession} and \textit{Tense} features, while no improvement has been seen for \textit{CharacterBin}, \textit{TagCount} or \textit{Pseudo}. Again, all tests with increasing scores had been shown to have significantly high correlation to Turkish SRL, while \textit{Pseudo} had no significant correlation. Other correlated features with non-increasing scores can be explained similar to the case for Finnish.
    
  \subsubsection{\rev{}{Token-Level Diagnosis}}
  \label{ssec:tok_diags}
\begin{figure*}[!ht]
      \centering
        \begin{subfigure}[b]{0.47\textwidth}
          \includegraphics[width=\textwidth]{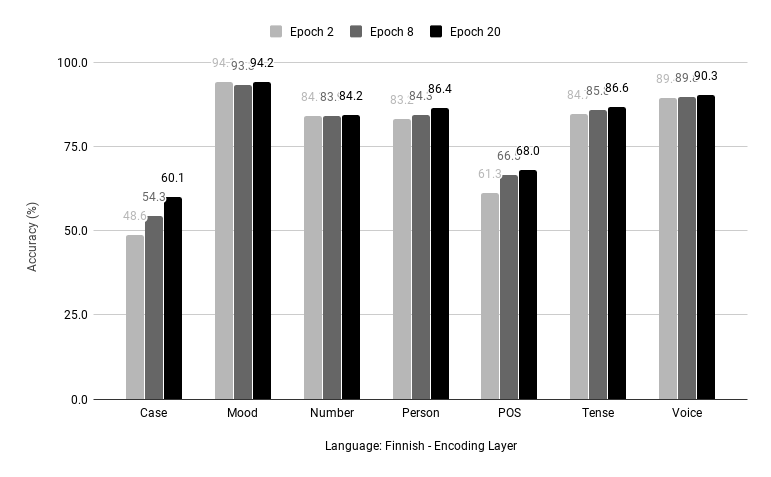}  
      \end{subfigure}
      \begin{subfigure}[b]{0.47\textwidth}
          \includegraphics[width=\textwidth]{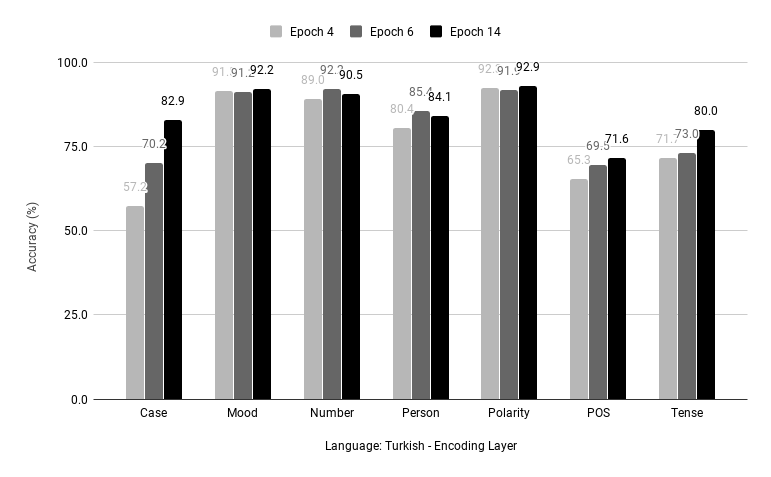} 
      \end{subfigure}
      
      \begin{subfigure}[b]{0.47\textwidth}
          \includegraphics[width=\textwidth]{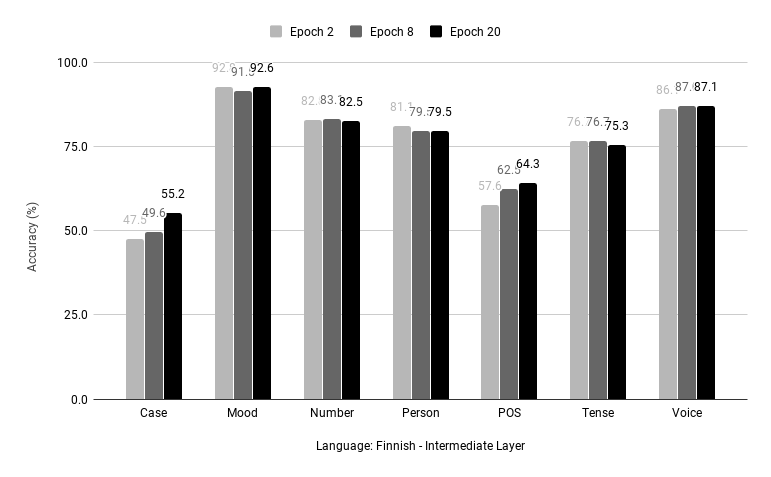}  
      \end{subfigure}
      \begin{subfigure}[b]{0.47\textwidth}
          \includegraphics[width=\textwidth]{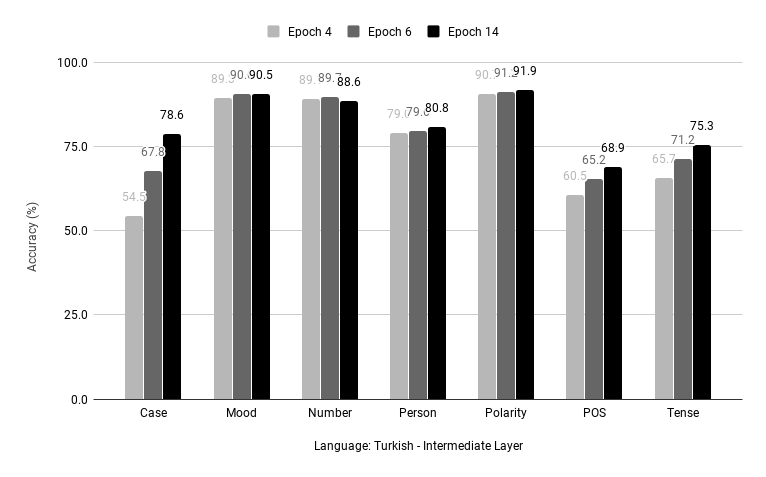} 
      \end{subfigure}
      \caption{\rev{}{Token-level probing tests to diagnose encoding and intermediate layers of pretrained Turkish and Finnish SRL models. \textbf{Top Left:} Finnish-Encoding; \textbf{Top Right:} Turkish-Encoding; \textbf{Bottom Left:} Finnish-Intermediate; \textbf{Bottom Right:} Turkish-Intermediate}}
      \label{fig:diagnostic_eval_token_encoding}
      \end{figure*}

       \rev{}{We show the results of our diagnostic case study with token-level probing tasks in Fig.~\ref{fig:diagnostic_eval_token_encoding}.}
     \rev{}{Probing of Finnish encoding layer shows that there is a significant performance improvement in \textit{Case} and \textit{POS} features, followed by \textit{Person}, \textit{Tense} and \textit{Voice} across subsequent epochs. Although the performance boost for \textit{Case} and \textit{POS} are still visible in intermediate layer results, we notice that \textit{Person} and \textit{Tense} features may have been forgotten in the next layer. This suggests that \textit{Case} and \textit{POS} features may be the crucial factors to predict semantic roles since they are conveyed to the next layer.} 
     
     \rev{}{We see a similar pattern for Turkish encoding layer, where there is a rise in accuracy scores of \textit{Case}, \textit{POS} and \textit{Tense}, followed by modest improvements in \textit{Person} and \textit{Mood} scores. Unlike Finnish that only transferred the \textit{Case} and \textit{POS} features to the next layer, Turkish seems to convey \textit{Person} and \textit{Tense} features along with \textit{Mood}. Although these features don't seem to be as crucial as \textit{Case} and \textit{POS} features that have been predominantly used as linguistic inputs to SRL systems, they may provide implicit clues. For instance, a predicate with first person singular tag is more likely to have an argument with an agent role; while a predicate with third person singular tag is less likely since most predicates in passive voice will be tagged with third person singular. Similarly, a predicate tagged with future tense, may be more likely to have a goal argument. In other words, these features are not mutually exclusive, and \textit{can not be} mutually exclusive.} 
     
\subsubsection{\rev{}{Discussion on Type- and Token-Level Diagnosis}}
\label{ssec:diags_compare}

    \rev{}{Next, we examine the word encoding diagnostic results for type and token-level tasks for comparison, starting with Finnish. First of all, almost all common features show the same increasing pattern for both languages hinting the similarity of both tasks, with exceptions on \textit{Mood} and \textit{POS} for Finnish; and \textit{Person} and \textit{Polarity} for Turkish. We hypothesized earlier that token-level tasks may be biased towards frequently occurring tag values. The difference in \textit{Mood} features may be a result of this observation. As discussed in Sec.~\ref{ssec:downs_findings}, \textit{Mood} can be a distinguishing feature for SRL. Since there are more sentences with ``Indicative'' predicates, \textit{Mood} feature in token-level probing will have a bias towards Indicative tag, resulting with inconclusive trajectory; while a more clear pattern is noticeable in type-level \textit{Mood} feature. Next, we see the opposite for \textit{POS} feature, where we have a clear pattern in token-level tasks, as opposed to a vague pattern in type-level tests. As we discussed earlier, this may be due to the limited lexical variety in type-level tasks unlike token-level tasks. When we move on with the Turkish results, we see the same tag frequency effect as in \textit{Mood} for Finnish pair, in \textit{Person} and \textit{Polarity} features. Even though \textit{POS} has a clearer pattern again in token-level diagnostics, the same pattern is still visible in type-level, suggesting that the lexical variety was not as severe as in Finnish case.} \rev{}{In order to encourage researchers to conduct similar multilingual diagnostic studies, we have also released a more convenient, online diagnostics platform that uses the proposed probing tasks~\cite{eichler2019linspector}.}

\subsection{\rev{Discussion on Probing Tests}{Summary of Experimental Findings}}
  Here we summarize our general findings on the proposed probing tasks from the experiments (see Sec.~\ref{ssec:results}) and analysis (see Sec.~\ref{ssect:correl}-\ref{ssec:diags}) sections.

  \begin{itemize}
  
  \item \rev{}{For all languages, the general ranking of the embeddings is: D-ELMo, fastText/GloVe-BPE, word2vec/MUSE for most \secrev{}{type-level} intrinsic tasks and the extrinsic tasks POS, DEP and SRL. \secrev{}{Although the same pattern is visible in token-level intrinsic experimental results, these embeddings are mostly surpassed by C-ELMo.}} 

  \item \rev{}{Static word embedding spaces (word2vec and MUSE) generally rank higher on downstream tasks compared to probing tasks, due to having lower OOV ratios in downstream tasks. }

  \item \rev{}{Low majority voting baseline scores in probing tasks generally mirror the morphological counting complexity (MCC) of the language (i.e., in case of more \emph{Case} categories, the lower the baseline). The trade-off between MCC and the morphological irregularity (i.e., the higher the MCC, the lower irregularity), later yields higher probing accuracies despite the low baselines.}

  \item \rev{}{Type-level probing tasks contain less domain and majority class bias, while the statistics of the resource, e.g., tag and token frequencies, have a direct impact on token-level tasks. However, token-level tests are generally more lexically diverse and more convenient to probe contextualized embedding models or intermediate hidden layers of black-box models. Removing ambiguous forms from a type-level probing task mostly results in weaker correlations as revealed by diagnostic and correlation studies; and mostly impacts the fusional languages where one morpheme may correspond to several morphological features.} \secrev{}{Despite these differences, the same ranking of various embeddings on both probing task types; and similar performance trajectories among epochs of SRL models for some common probing tasks hint at the commonalities between type and token level tasks.}


  \item The calculated correlations were positive.
  
  \item The set of correlated tests were generally small for fusional languages, and large for agglutinative languages.
  
  \item The set of correlated tests varied with the \emph{complexity} of the downstream task, however the correlation pattern was common across similar tasks \textit{(e.g., SRL had a large set of correlated tests, while POS tagging has a only subset of it)}.
  
  \item The set of correlated tests varied with the \emph{requirements} of the downstream task (e.g., paired tests like \textit{SameFeat} had strong correlation to XNLI, but had weak ties to syntactic tasks).
  
  \item We observe commonalities among the correlated probing tests for Finnish, Turkish, German and Russian. For instance the correlation between \textit{Case}, \textit{POS}, \textit{Person}, \textit{Tense}, \textit{TagCount} and the downstream tasks were higher than the other probing tests. This suggests that the findings are transferable, hence the proposed probing tests can be used for other languages.
  
  \item We also observe that language specific tests are beneficial, i.e., have significantly high correlation such as \textit{Polarity} for Turkish, and some tests could be impactful for a language family, e.g. \textit{CharacterBin} for agglutinative languages, \textit{Pseudo} for Spanish.

  \item \rev{}{For almost all languages, there is a lack of correlation between the probing tests and NER performance, which we attribute to the shallow, grammar-agnostic nature of the NER task}.

  \item \rev{}{Apart from linguistic properties, dataset statistics play a crucial role on the results (e.g., domain similarity for Finnish; lexical variety for Spanish; low OOV ratio for all XNLI tasks) and may add noise to correlation study. When number of data points is not enough to reliably estimate the correlation, the noise should be interpreted carefully.}


  \item \rev{}{There is a strong connection between the correlated tests and the morphological features captured throughout the epochs of black-box Finnish and Turkish SRL models, suggesting that diagnostics can be a useful application of the probing tasks.}

  \end{itemize}

  To follow up on our discussion on strong baselines, and the difficulty of the tests from Sec.~\ref{ssec:results}, we find that correlations neither depend on how strong the initial baseline is, nor how low the accuracy scores for this test are. For instance, Part-of-Speech tagging has strong baselines for many languages, however it is also one of our most correlated tests. Moreover, ``hard'' tests with low scores, such as \textit{OddFeat} and \textit{CharacterBin} behaved like any other tests, i.e., had low-to-high levels of correlation for different language/downstream task pairs. 



\section{Conclusion}
In this study we have introduced 15 type-level probing tests for a total of 24 languages, where the target linguistic phenomena differ depending on the typological properties of the language, e.g., \textit{Case}, \textit{Polarity} for Turkish, \textit{Gender} for Russian and German. These tests are proposed as an exploratory tool to reveal the underlying linguistic properties captured by a word embedding or a layer of a neural model trained for a downstream task. Furthermore, we introduce a methodology for creation and evaluation of such tests which can be easily extended to other datasets and languages. We release the framework LINSPECTOR with \url{https://github.com/UKPLab/linspector}, that consists of the datasets for probing tasks along with an easy-to-use probing and downstream evaluation suite based on AllenNLP.

We have performed an exhaustive set of intrinsic and extrinsic experiments with \rev{commonly used}{a diverse set of} pretrained multilingual embeddings for five typologically diverse languages: German, Spanish, Russian, Turkish and Finnish. We found that evaluated embeddings provide a varying range of improvement over the baselines. 
Our statistical analysis of intrinsic and extrinsic experimental results showed that the proposed probing tasks are positively correlated to majority of the downstream tasks. In general, the number of correlated probing tests was higher for agglutinative languages, especially for syntactic tasks. We showed that the sets of correlated tests differ depending on the type of the downstream task. For instance XNLI performance is strongly correlated with the \textit{SameFeat} probing accuracy, while SRL is correlated well with the \textit{Case}. 
We observed \textit{Case}, \textit{POS}, \textit{Person}, \textit{Tense} and \textit{TagCount} to have significantly high correlations for majority of the analyzed languages and tasks; in addition, language specific tests such as \textit{Possession} were found to correlate well in cases when they were applicable. Furthermore, the results of our diagnostic case study, where we probe encoding \rev{}{and an intermediate} layer of a black-box neural model, showed strong connections to the correlated tests. All these findings suggest that the proposed probing tests can be used to estimate the predictive performance of an input representation on a downstream task, as well as to explore the strengths and weaknesses of existing neural models, or to understand the relation between a model parametrization and its ability to capture linguistic information, \textit{ e.g., how the performance on probing tests changes after increasing the model size)}.  

\rev{}{We have shown that dataset statistics \emph{(e.g., out-of-vocabulary ratio, dataset size)} are a major factor influencing the results in addition to linguistic properties of the language \emph{(e.g., typology, paradigm size, regularity)}. This can sometimes introduce noise and yield to inconclusive correlations. Finally, investigation of \textbf{token-level} tasks revealed that \textbf{type-level} tasks contain less domain and majority class bias compared to token-level tasks; while token-level tests are generally more lexically diverse. In addition, removal of ambiguous forms from a type-level probing task may result in weaker correlations.} 

\rev{As future work, we plan to extend our evaluation suite. Current \ikrev{evaluation suite }{implementation} only supports probing of static word embeddings, which is not convenient for exploring black box models. Hence we plan to implement a feature for existing AllenNLP models, where the user can specify the layer and the language to be probed, and obtain the probing results automatically. Furthermore, the methodology we introduced can be used to create ``contextual'' probing tests with a small effort and to explore the properties of contextualized embeddings by using the recently introduced SIGMORPHON 2019 dataset adapted from Universal Dependencies.}{}

\section{Acknowledgements}

We first thank to anonymous reviewers that helped us improve the paper. We would like to thank Marvin Kaster for his help on contextualizing the probing tasks and to Max Eichler for his contribution on acquiring experimental results for additional languages in Appendix L. We are sincerely grateful to Adam Lopez, Ida Szubert, Ji-Ung Lee, Edwin Simpson and members of AGORA Lab for providing feedback on early drafts of this work. This research has been supported by the DFG-funded research training group ``Adaptive Preparation of Information form Heterogeneous Sources'' (AIPHES, GRK 1994/1), and also by the German Federal Ministry of Education and Research (BMBF) under the promotional reference 01UG1816B (CEDIFOR) and as part of the Software Campus program under the promotional reference 01IS17050. We gratefully acknowledge the support of NVIDIA Corp. with the donation of the Tesla K40 GPU used for this research. Clara Vania is supported by the Indonesian Endowment Fund for Education (LPDP), the Centre for Doctoral Training in Data Science, funded by the UK EPSRC (grant EP/L016427/1), and the University of Edinburgh. We finally thank to UKP system administrators for their system maintenance efforts and Celal Şahin for helping with system configuration errors. 

\clearpage

\section{Appendix}

\appendixsection{\rev{}{Out-of-Vocabulary Analysis}}
\label{app:oov-rate}

\begin{table}[ht]
    \centering
    \footnotesize
    \caption{\rev{}{Training and development OOV rate for intrinsic task with MUSE}}
    \label{tab:muse-intrinsic-oov}
    \begin{tabular}{lcccccccccc}
    \toprule
    \multirow{2}{*}{Feature} & \multicolumn{2}{c}{Finnish} & \multicolumn{2}{c}{German} & \multicolumn{2}{c}{Russian} & \multicolumn{2}{c}{Spannish} & \multicolumn{2}{c}{Turkish} \\
    \cmidrule{2-11}
    & train & dev & train & dev & train & dev & train & dev & train & dev \\
    \midrule
    Case & 61.9 & 59.7 & 56.1 & 56.5 & 61.7 & 62.8 & - & - & 44.8 & 44.9 \\
    Gender & - & - & - & - & 63.4 & 64.2 & 47.9 & 48.8 & - & - \\
    Mood & 55.6 & 56.6 & 78.0 & 76.3 & - & - & 51.7 & 49.7 & - & - \\
    Number & 59.7 & 58.6 & 59.7 & 60.3 & 64.5 & 66.2 & 49.6 & 50.5 & - & - \\
    POS & 71.5 & 71.6 & 72.1 & 70.2 & 73.7 & 71.9 & 62.7 & 63.3 & 46.0 & 44.7 \\
    Person & 55.4 & 58.0 & 78.4 & 78.0 & 67.4 & 67.3 & 51.1 & 52.5 & 57.3 & 56.1 \\
    Polarity & - & - & - & - & - & - & - & - & 56.8 & 56.4 \\
    Possession & - & - & - & - & - & - & - & - & 45.3 & 45.5 \\
    Pseudo & - & - & 46.5 & 49.2 & - & - & 50.8 & 50.6 & 52.2 & 51.0 \\
    Tense & 55.2 & 55.6 & 72.4 & 72.8 & 67.5 & 68.3 & 50.8 & 49.0 & 50.8 & 52.5 \\
    Voice & 54.8 & 54.6 & - & - & 83.0 & 82.2 & - & - & - & - \\
    \midrule
    CharacterBin & 67.9 & 68.4 & 59.7 & 58.0 & 68.0 & 68.2 & 63.9 & 63.8 & 44.1 & 44.3 \\
    TagCount & 67.9 & 68.4 & 59.7 & 58.0 & 68.0 & 68.2 & 63.9 & 63.8 & 44.1 & 44.3 \\
    \midrule
    OddFeat & 87.2 & 88.7 & 50.9 & 49.4 & 60.3 & 60.0 & 64.3 & 65.2 & 62.2 & 61.0 \\
    SameFeat & 91.7 & 92.2 & 50.7 & 49.7 & 65.6 & 64.8 & 83.4 & 83.8 & 78.7 & 78.4 \\
    \bottomrule
    \end{tabular}
\end{table}

\begin{table}[ht]
    \centering
    \footnotesize
    \caption{\rev{}{Training and development OOV rate for intrinsic task with word2vec}}
    \label{tab:word2vec-intrinsic-oov}
    \begin{tabular}{lcccccccccc}
    \toprule
    \multirow{2}{*}{Feature} & \multicolumn{2}{c}{Finnish} & \multicolumn{2}{c}{German} & \multicolumn{2}{c}{Russian} & \multicolumn{2}{c}{Spannish} & \multicolumn{2}{c}{Turkish} \\
    \cmidrule{2-11}
    & train & dev & train & dev & train & dev & train & dev & train & dev \\
    \midrule
    Case & 19.7 & 18.6 & 34.5 & 33.5 & 18.6 & 18.6 & - & - & 21.7 & 20.1 \\
    Gender & - & - & - & - & 18.3 & 17.9 & 22.3 & 21.4 & - & - \\
    Mood & 19.4 & 20.4 & 66.7 & 65.5 & - & - & 38.6 & 37.9 & - & - \\
    Number & 19.4 & 18.9 & 38.2 & 37.1 & 18.7 & 18.2 & 31.2 & 30.9 & - & - \\
    POS & 20.3 & 19.8 & 45.5 & 43.4 & 36.3 & 36.6 & 37.8 & 37.4 & 21.8 & 19.9 \\
    Person & 19.2 & 20.6 & 66.5 & 68.1 & 19.1 & 19.9 & 39.0 & 38.7 & 35.6 & 35.4 \\
    Polarity & - & - & - & - & - & - & - & - & 35.5 & 33.8 \\
    Possession & - & - & - & - & - & - & - & - & 21.3 & 21.9 \\
    Pseudo & - & - & 51.6 & 52.1 & - & - & 42.1 & 42.0 & 47.9 & 47.5 \\
    Tense & 20.0 & 17.9 & 59.5 & 58.0 & 18.7 & 20.0 & 37.8 & 35.6 & 27.6 & 27.8 \\
    Voice & 18.9 & 20.0 & - & - & 54.6 & 51.5 & - & - & - & - \\
    \midrule
    CharacterBin & 20.3 & 21.4 & 39.9 & 41.0 & 33.9 & 34.0 & 35.7 & 36.0 & 21.1 & 21.5 \\
    TagCount & 20.3 & 21.4 & 39.9 & 41.0 & 33.9 & 34.0 & 35.7 & 36.0 & 21.1 & 21.5 \\
    \midrule
    OddFeat & 75.7 & 77.2 & 38.4 & 39.5 & 29.7 & 29.6 & 46.3 & 44.8 & 49.3 & 50.2 \\
    SameFeat & 85.1 & 85.9 & 38.7 & 38.3 & 37.0 & 38.0 & 73.1 & 73.5 & 70.6 & 71.0 \\
    \bottomrule
    \end{tabular}
\end{table}

\begin{table}[ht]
    \centering
    \footnotesize
    \caption{\rev{}{OOV rate for extrinsic task with MUSE and word2vec embeddings}}
    \label{tab:muse-word2vec-extrinsic-oov}
    \begin{tabular}{llrrrrrr}
    \toprule
    \multirow{2}{*}{Task} & \multirow{2}{*}{Language} & \multicolumn{3}{c}{MUSE} & \multicolumn{3}{c}{word2vec} \\
    \cmidrule{3-8}
    & & train & dev & test & train & dev & test \\
    \midrule
    NER & Finnish & 16.33 & 14.44 & 16.34 & 11.28 & 9.07 & 11.21 \\
    & German & 9.28 & 9.28 & 9.19 & 14.02 & 13.81 & 14.18 \\
    & Russian & 11.73 & 11.54 & 11.92 & 8.8 & 8.62 & 8.97 \\
    & Spanish & 4.34 & 5.13 & 3.89 & 13.63 & 13.88 & 13.14 \\
    & Turkish & 13.77 & 13.69 & 13.6 & 12.24 & 12.17 & 11.96 \\
    \midrule
    UD & Finnish & 15.86 & 15.17 & 15.08 & 10.68 & 10.32 & 10.02 \\
    & German & 7.86 & 7.05 & 7.61 & 11.74 & 11.55 & 12.36 \\
    & Russian & 7.03 & 7.26 & 7.88 & 3.41 & 3.69 & 4.26 \\
    & Spanish & 2.71 & 2.79 & 2.98 & 12.08 & 11.85 & 12.32 \\
    & Turkish & 10.66 & 10.53 & 10.36 & 7.79 & 7.98 & 7.87 \\
    \midrule
    XNLI & German & 3.22 & 5.21 & 5.47 & 9.4 & 10.58 & 10.62 \\
    & Russian & 4.25 & 7.19 & 7.15 & 2.81 & 3.95 & 4.02 \\
    & Spanish & 1.95 & 2.79 & 3.01 & 10.98 & 11.89 & 11.93 \\
    & Turkish & 3.98 & 8.13 & 8.39 & 2.99 & 5.77 & 5.89 \\
    \midrule
    SRL & Finnish & 15.92 & 14.82 & 15.69 & 10.75 & 9.92 & 10.9 \\
    & German & 9.89 & 10.83 & 11.12 & 14.68 & 15.39 & 15.48 \\
    & Spanish & 5.86 & 5.96 & 6.16 & 14.72 & 14.56 & 15.01 \\
    & Turkish & 10.25 & 9.9 & 9.92 & 7.58 & 7.28 & 7.26 \\
    \bottomrule
    \end{tabular}
\end{table}

\begin{figure}[!ht]
  \centering
    \begin{subfigure}[b]{0.45\textwidth}
      \includegraphics[width=\textwidth]{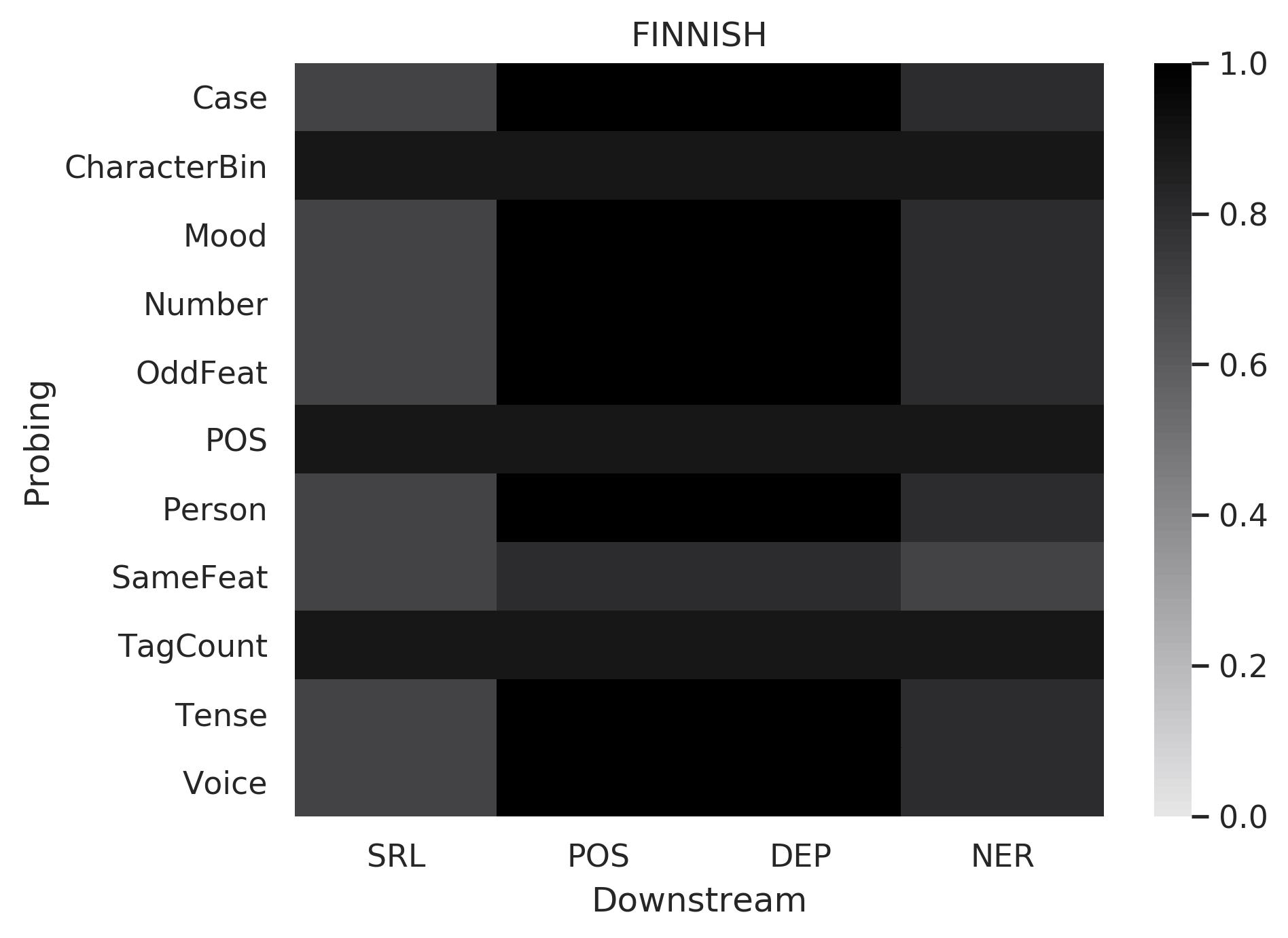}  
  \end{subfigure}
  ~ 
  \begin{subfigure}[b]{0.45\textwidth}
      \includegraphics[width=\textwidth]{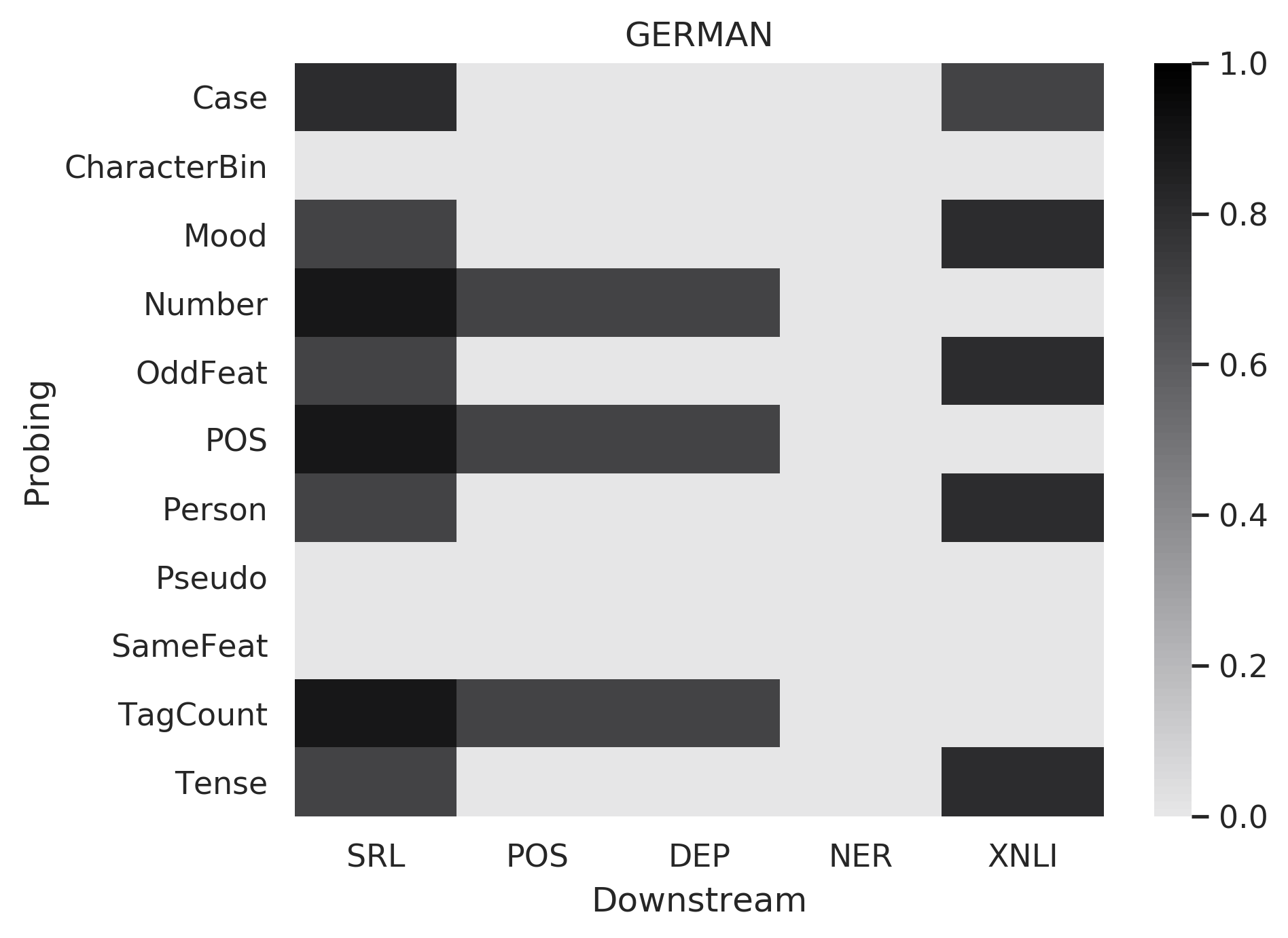} 
  \end{subfigure}
  ~ 
  \begin{subfigure}[b]{0.45\textwidth}
      \includegraphics[width=\textwidth]{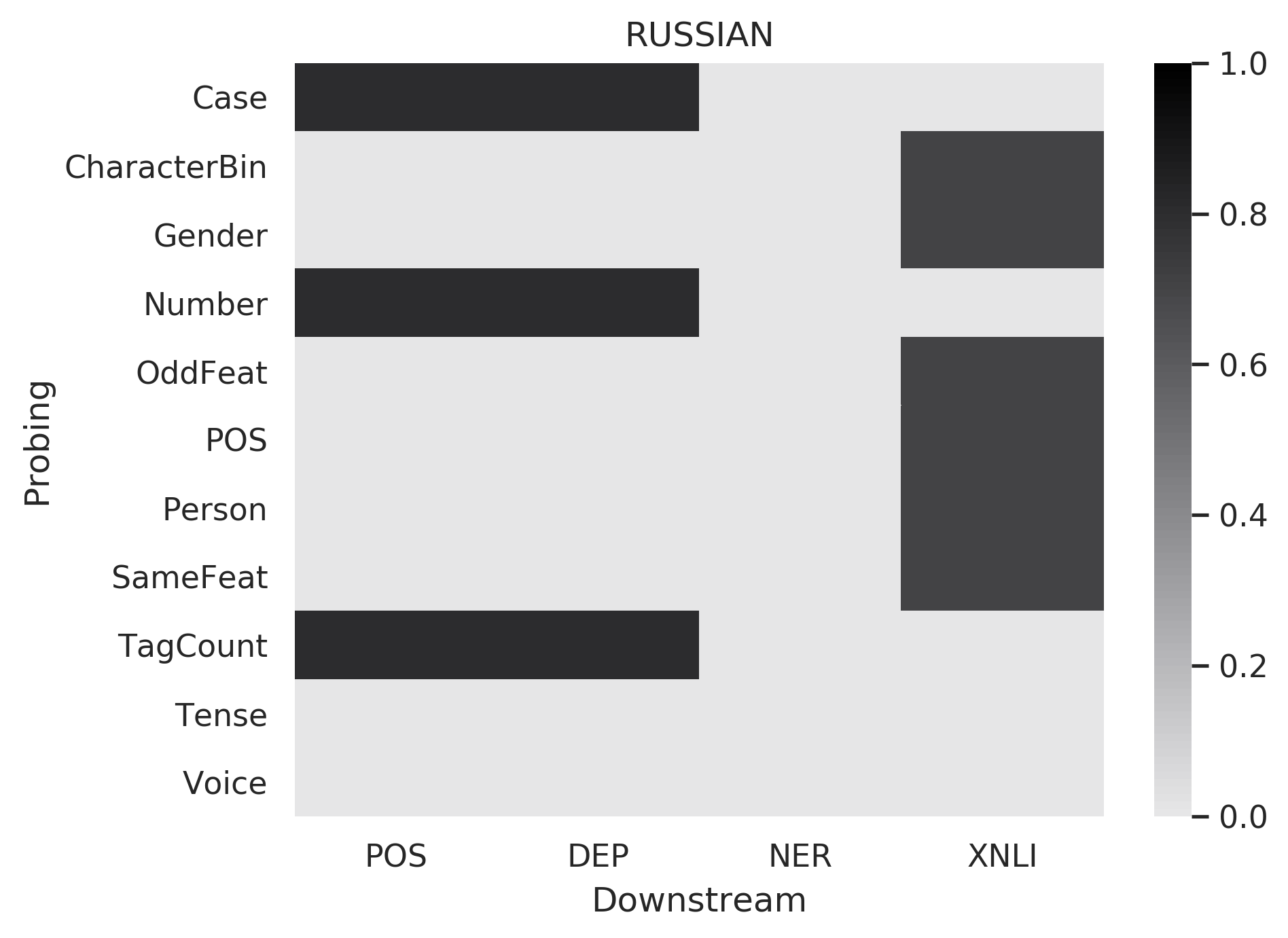}
  \end{subfigure}
  ~ 
  \begin{subfigure}[b]{0.45\textwidth}
      \includegraphics[width=\textwidth]{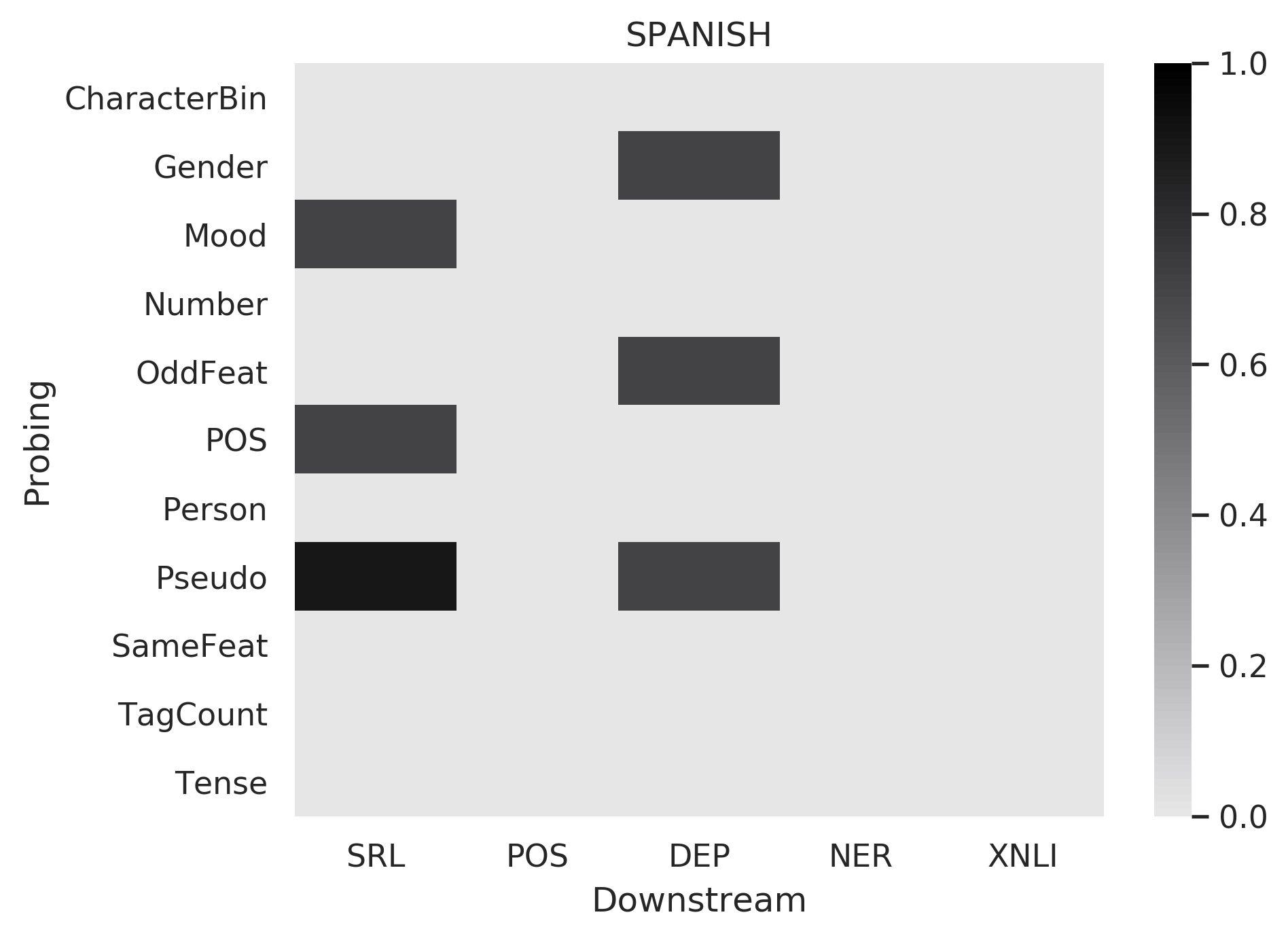}
  \end{subfigure}
  ~ 
  \begin{subfigure}[b]{0.45\textwidth}
      \includegraphics[width=\textwidth]{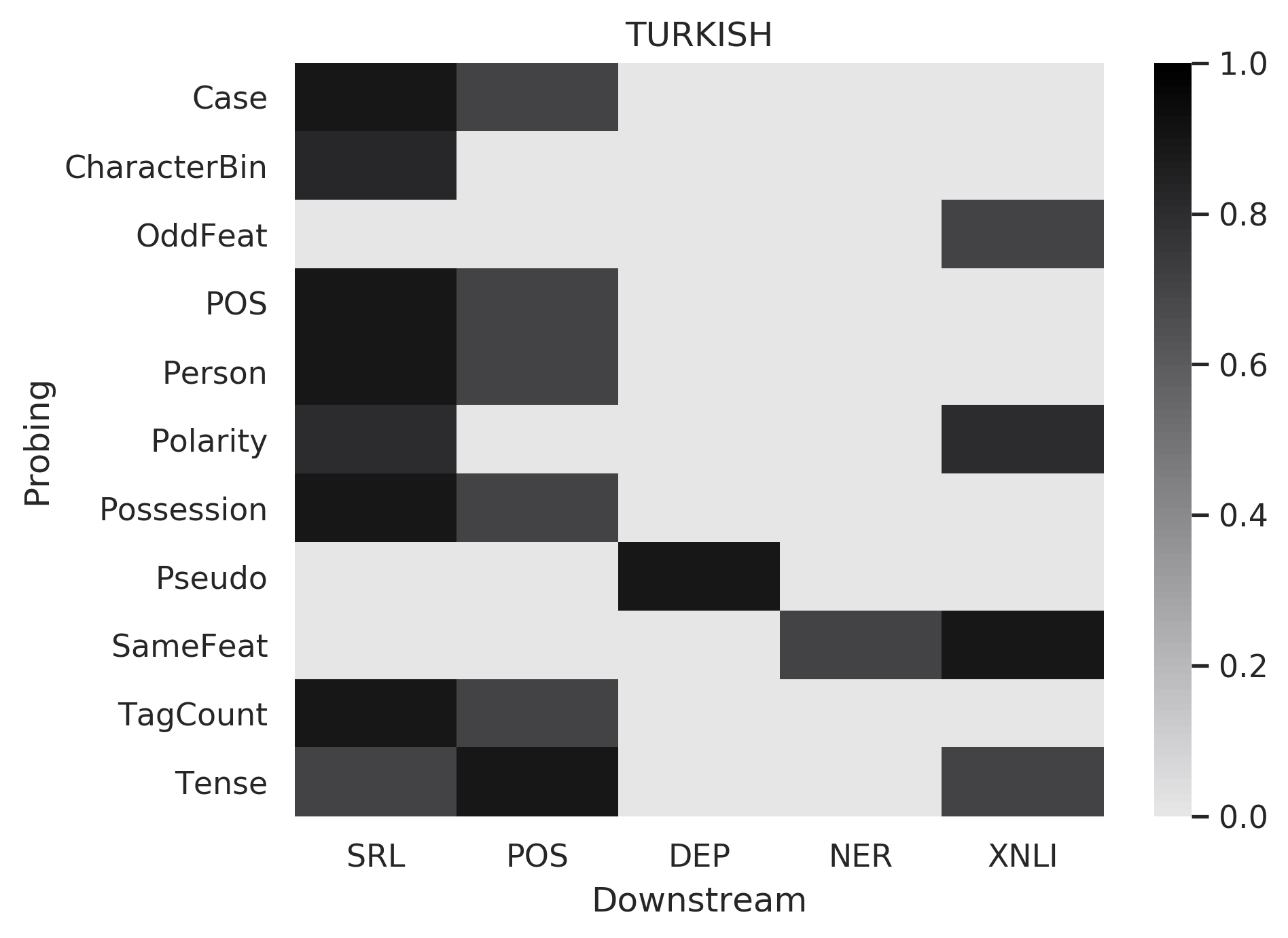}
  \end{subfigure}
  \caption{Spearman correlation between probing and downstream tasks for each language. Weak correlations are not shown.}
  \label{fig:spearman_sign}
  \end{figure}

\clearpage
\appendixsection{Correlation}
\label{app:sec1}
In order to provide more insight for the relation between the downstream tasks and the probing tests, we show only the significant Spearman correlations with $p=0.2$ in Fig.~\ref{fig:spearman_sign}.

\clearpage
\appendixsection{\rev{}{Intrinsic Experiments on Additional Languages}}
\label{app:sec2}

\begin{figure}[!ht]
\centering
    \begin{subfigure}[b]{0.45\textwidth}
      \includegraphics[width=\textwidth]{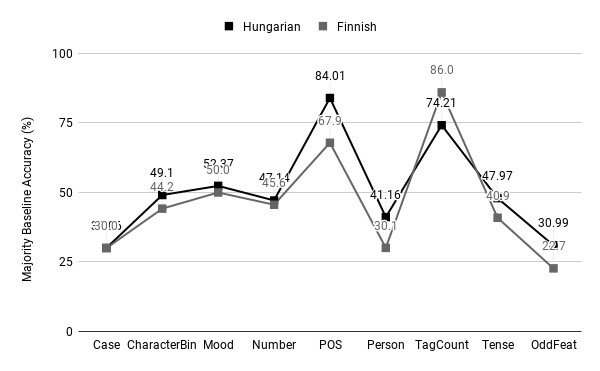}  
    \end{subfigure}
    ~
    \begin{subfigure}[b]{0.45\textwidth}
      \includegraphics[width=\textwidth]{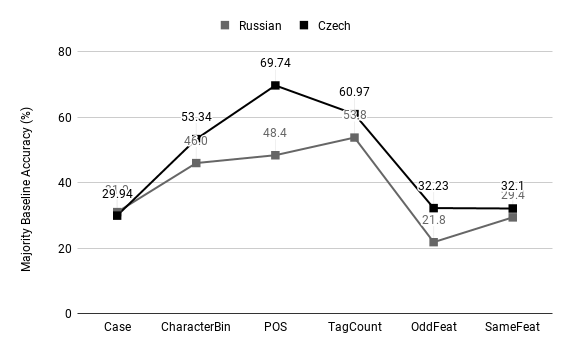} 
    \end{subfigure}

    \begin{subfigure}[b]{0.45\textwidth}
      \includegraphics[width=\textwidth]{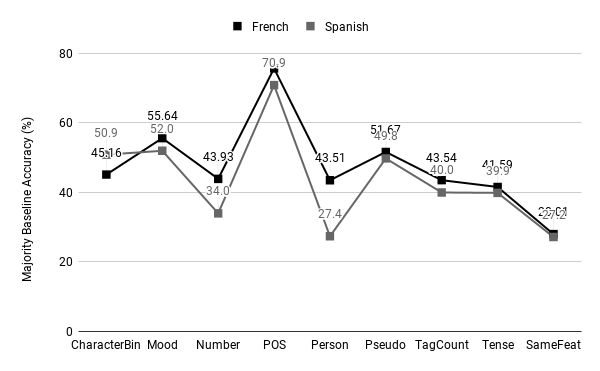} 
    \end{subfigure}
\caption{\rev{}{Majority baseline scores comparison for related languages for the common tasks}}
\label{fig:rel_langs}
\end{figure}

\rev{}{Our framework allows to compare the type-level probing task performance between related languages, and we report additional results on Czech from Slavic; Hungarian from Uralic; and French from Romance language families. We first compare the majority baseline scores for each language pair shown in Fig.~\ref{fig:rel_langs}, then apply the same intrinsic experimental setup for the additional languages given in Table~\ref{tab:other-probing-results}.} 

\rev{}{\paragraph{Hungarian vs Finnish} As it can be seen in Fig.~\ref{fig:rel_langs} Hungarian and Finnish follow similar performance trends, in line with the similar morphological counting complexity (MCC) scores as given by~\citep{CotterellMER18}. The slightly bigger gap between POS tasks is due to Finnish data containing words from the ``Adjective'' class, which Hungarian data doesn't. When compared, the ranking of different embedding spaces for intrinsic experiments were identical for both languages: D-ELMO, followed by fastText, which was generally true for all languages experimented with. In addition, we observe that while all embeddings achieved relatively high scores for ``Number", ``POS" and ``TagCount" probing tests, ``Tense", ``CharacterBin" and ``OddFeat" had lower scores for both languages. Apart from similarities, when we compare the scores of the best performing embedding on intrinsic experiments, we observe that it achieved lower scores for the majority of the tasks for Finnish. This can be explained with high number of paradigms (57642) in Finnish Unimorph data, compared to Hungarian data that only has 13989 paradigms.} 

\rev{}{\paragraph{Czech vs Russian} While the majority baseline scores follow the same trajectory for both languages, we observe the repeating pattern of lower baseline scores for probing tasks in Russian with respect to Czech. The only exception is the ``Case" probing task, which can be explained by the different numbers of case markers: Czech has 7, while Russian has 6 distinct case marking features in the Unimorph data. These results suggest that Russian probing tasks had more heterogeneous instances (especially for the POS test). This is particularly due to the original Czech Unimorph data covering predominantly nominal inflections while Russian data is more balanced in terms of lexical variety. This is hinted by the large gap between the Unimorph datasizes, where Russian had 28068 paradigms and 473481 inflections, while Czech data was less than a quarter of it (5125 and 134527, respectively). 
Finally, we find that the accuracy intrinsic task ranking of the best performing embedding, D-ELMO, is identical across both languages, i.e., ``POS" having the highest scores while ``OddFeat" has the lowest.} 

\rev{}{\paragraph{French vs Spanish} Unlike previous language pairs, Unimorph 2.0 only contains verbal inflection paradigms for both languages, which slightly limits our analysis. Similar to previous language pairs, we observe a comparable baseline trajectory as shown in Fig.~\ref{fig:rel_langs}, with a slightly bigger gap for Number and Person tasks. We attribute it to the removal of more ambiguous forms or infrequent words that contain a diverse set of Number and Person tags in French than Spanish. When comparing the results of the intrinsic experiments, we notice a repeating pattern where all embeddings score their highest for Number, POS and Person; while achieving the lowest for CharacterBin and SameFeat tasks.} 

\begin{table}[!ht]
    \centering
    \caption{\rev{}{Probing task results for additional languages from the same language family. \textbf{Bold} represents the best score, while \textit{italics} is the second best.}}
    \label{tab:other-probing-results}
    \scalebox{0.65}{
    \begin{tabular}{lcccccc}
    \hline
    \multicolumn{7}{c}{\textit{Hungarian (Uralic)}} \\
    \hline
    \toprule
    Task & baseline & MUSE & word2vec & GloVe-BPE & fastText & D-ELMo \\
    \midrule
    Case & 30.06 & 55.7 & 43.2 & 86.8 & \textit{95} & \textbf{98.2} \\
    Mood & 52.37 & 64.3 & 68.7 & 83.9 & \textit{94.3} & \textbf{97.6} \\
    Number & 47.14 & 63.3 & 61 & 85.9 & \textit{94.8} & \textbf{97.3} \\
    POS & 84.01 & 89 & 85 & 92.2 & \textit{97.8} & \textbf{99.6} \\
    Person & 41.16 & 58.6 & 58.1 & 82.9 & \textit{93.8} & \textbf{97} \\
    Tense & 47.97 & 67.1 & 65.9 & 81.3 & \textit{93.2} & \textbf{95.1} \\
    \midrule
    CharacterBin & 49.1 & 53.9 & 47.6 & 52 & \textit{55.4} & \textbf{65} \\
    TagCount & 74.21 & 83.7 & 77.1 & 90.3 & \textit{96.4} & \textbf{98.2} \\
    \midrule
    OddFeat & 30.99 & 37.5 & 32.5 & 45.8 & \textit{49.2} & \textbf{52.2} \\
    \bottomrule
    \hline
    \multicolumn{7}{c}{\textit{Czech (Slavic)}} \\
    \hline
    \toprule
    Task & baseline & MUSE & word2vec & GloVe-BPE & fastText & D-ELMo \\
    \midrule
    Case & 29.94 & 69.6 & 64.4 & 85.8 & \textit{94.6} & \textbf{97.1} \\
    POS & 69.74 & 80.3 & 72.2 & 87.6 & \textit{96.1} & \textbf{98.5} \\
    \midrule
    CharacterBin & 53.34 & 57.5 & 54.2 & 61.0 & \textit{65.8} & \textbf{71.5} \\
    TagCount & 60.97 & 78.2 & 68.3 & 78.2 & \textit{92.2} & \textbf{95.8} \\
    \midrule
    OddFeat & 32.23 & 46 & 42.4 & 41.7 & \textit{49.7} & \textbf{52.2} \\
    SameFeat & 32.1 & 65.9 & 69.6 & 75.7 & \textit{84} & \textbf{89.7} \\
    \bottomrule
    \hline
    \multicolumn{7}{c}{\textit{French (Romance)}} \\
    \hline
    \toprule
    Task & baseline & MUSE & word2vec & GloVe-BPE & fastText & D-ELMo \\
    \midrule
    Mood & 55.64 & 70.6 & 72.2 & 79.2 & \textit{93.4} & \textbf{93.4} \\
    Number & 43.93 & 62.9 & 72.1 & 88 & \textit{98.3} & \textbf{99} \\
    POS & 75.7 & 88.7 & 84.6 & 91.4 & \textit{98.8} & \textbf{99.3} \\
    Person & 43.51 & 58 & 66.4 & 85.7 & \textit{97.7} & \textbf{98} \\
    Pseudo & 51.67 & \textit{96.9} & 75 & 90.8 & 85.5 & \textbf{97.6} \\
    Tense & 41.59 & 62.7 & 67.6 & 73.5 & \textbf{95.3} & \textit{94} \\
    \midrule
    CharacterBin & 45.16 & 57.6 & 54.6 & 59.5 & \textit{62.1} & \textbf{73.9} \\
    TagCount & 43.54 & 66.9 & 67 & 67.9 & \textbf{91.1} & \textit{90.2} \\
    \midrule
    SameFeat & 28.01 & 43.2 & 67.3 & 55.3 & \textit{72.2} & \textbf{72.6} \\
    \bottomrule
    \end{tabular}
    }
\end{table}

\clearpage
\starttwocolumn
\bibliography{compling_style}

\end{document}